%% file: main.tex
\documentclass[10pt,twocolumn,letterpaper]{article}

\usepackage{cvpr}    
\usepackage{multirow}
\usepackage{booktabs}
\usepackage[table]{xcolor} 
\newcommand{\faintmidrule}{\arrayrulecolor{black!15}\midrule\arrayrulecolor{black}}
\definecolor{cvprblue}{rgb}{0.21,0.49,0.74}
\usepackage[pagebackref,breaklinks,colorlinks,allcolors=cvprblue]{hyperref}
\usepackage{graphicx,makecell,rotating,booktabs}
\definecolor{LightGray}{gray}{0.9}
\definecolor{mygray}{gray}{0.9}

\usepackage[ruled,vlined]{algorithm2e}
\usepackage{pifont} 

\usepackage{tikz}
\usetikzlibrary{arrows.meta,positioning,fit,calc,decorations.markings,backgrounds}

\tikzset{
  >={Latex[length=3mm]},
  module/.style      ={draw, rounded corners=2pt, minimum height=8mm, minimum width=16mm, align=center, font=\small, fill=white},
  frozen/.style      ={module, dashed, very thick, draw=blue!70, fill=blue!5},
  trainable/.style   ={module, ultra thick, draw=orange!70!black, fill=orange!6},
  pool/.style        ={module, draw=black!70, fill=gray!6},
  loss/.style        ={module, draw=red!50!black, fill=red!3},
  note/.style        ={font=\scriptsize, inner sep=1pt, text=black!70},
  group/.style       ={draw=black!40, rounded corners=4pt},
  fw/.style          ={-Latex, line width=0.9pt},                 
  bp/.style          ={-Latex, dashed, line width=0.9pt, draw=red!70}, 
  sgmark/.style      ={circle, draw=red!70, fill=white, inner sep=0.4pt, minimum size=3.6mm, font=\scriptsize\bfseries, text=red!80},
  frozmark/.style    ={rectangle, draw=blue!70, fill=white, inner sep=0.4pt, font=\scriptsize, text=blue!80},
}

\input{preamble}
\def\paperID{2169} 
\def\confName{CVPR}
\def\confYear{2026}

\title{FairLLaVA: Fairness-Aware Parameter-Efficient Fine-Tuning for Large Vision-Language Assistants}

\author{
Mahesh Bhosale \textsuperscript{1}
\and
Abdul Wasi \textsuperscript{1}\thanks{Abdul Wasi and Shantam Srivastava contributed equally as joint second authors.}
\and
Shantam Srivastava \textsuperscript{1}\footnotemark[1]
\and
Shifa Latif \textsuperscript{2}
\and
Tianyu Luan \textsuperscript{3}
\and
Mingchen Gao \textsuperscript{1}
\and
David Doermann \textsuperscript{1}
\and
Xuan Gong \textsuperscript{4}
\\
\\
\textsuperscript{1}University at Buffalo
\quad
\textsuperscript{2}University of Kashmir
\quad
\textsuperscript{3}Accenture
\quad
\textsuperscript{4}Harvard Medical School
}
\begin{document}
\maketitle
\input{sec/0_abstract}
\input{sec/1_intro}
\input{sec/2_related_work}
\input{sec/4_Methods}

\input{sec/5_experiments}

\input{X_suppl}
{
    \small
    \bibliographystyle{ieeenat_fullname}
    \bibliography{CVPR-2026/author-kit-CVPR2026-v1-latex-/main}
}
\end{document}

%% file: sec/0_abstract.tex
\begin{abstract}
While powerful in image-conditioned generation, multimodal large language models (MLLMs) can display uneven performance across demographic groups, highlighting fairness risks. In safety-critical clinical settings, such disparities risk producing unequal diagnostic narratives and eroding trust in AI-assisted decision-making. While fairness has been studied extensively in vision-only and language-only models, its impact on MLLMs remains largely underexplored.
To address these biases, we introduce FairLLaVA, a parameter-efficient fine-tuning method that mitigates group disparities in visual instruction tuning without compromising overall performance.
By minimizing the mutual information between target attributes, FairLLaVA regularizes the model’s representations to be demographic-invariant. 
The method can be incorporated as a lightweight plug-in, maintaining efficiency with low-rank adapter fine-tuning, and provides an architecture-agnostic approach to fair visual instruction following.
Extensive experiments on large-scale chest radiology report generation and dermoscopy visual question answering benchmarks show that FairLLaVA consistently reduces inter-group disparities while improving both equity-scaled clinical performance and natural language generation quality across diverse medical imaging modalities. Code can be accessed at \href{https://github.com/bhosalems/FairLLaVA}{https://github.com/bhosalems/FairLLaVA}.
\end{abstract}

%% file: sec/1_intro.tex
\section{Introduction}
\label{sec:intro}
Large language models (LLMs) have emerged as general-purpose engines for reasoning and text generation across many domains, driving a wave of \emph{foundation model} capabilities at unprecedented scale \citep{bommasani2021opportunities,Radford2019LanguageMA,openai2023gpt4}. Multi-modal large language models (MLLMs) build on these advances by pairing a language backbone with non-text encoders (e.g., images), enabling image-grounded understanding and generation\citep{alayrac2022flamingo,yang2025qwen3, sellergren2025medgemma, chen2024chexagent, llavarad, llava, guo2025deepseek}. Despite these advances, a growing body of work documents disparities and fairness issues in these models \citep{cares2024,nature_evaluating, fairbench}. 

In natural-language domains, societal biases can enter when training datasets are not carefully curated or filtered~\cite{toward_aistat}. For example, models often reproduce gender stereotypes in professions, such as more strongly associating “nurse” with women and “engineer” with men~\cite{dear_CVPR, toward_aistat, t2i_fair}. However, in safety-critical medical settings, sensitive attributes are rarely explicitly mentioned in clinical reports. In particular, fairness considerations in radiology report generation (RRG)-an automated task that translates medical images into expert-quality textual interpretations~\cite{yu2023progress,rate2024,llavarad}-differ fundamentally from pronoun- or sentiment-based notions in the general language domain. Radiology reports seldom contain explicit demographic markers (e.g., “he/she”), yet models can still exploit latent sensitive information embedded in medical images and acquisition pipelines~\cite{gichoya2022race,lottter2024acq}. This renders standard word-level (polarizing-term) probes~\cite{llm_fair_1,llm_fair_2,llm_fair_3,toward_aistat} based on demographic tokens largely inapplicable.

\begin{figure}[t]
        \centering
        \includegraphics[width=\linewidth]{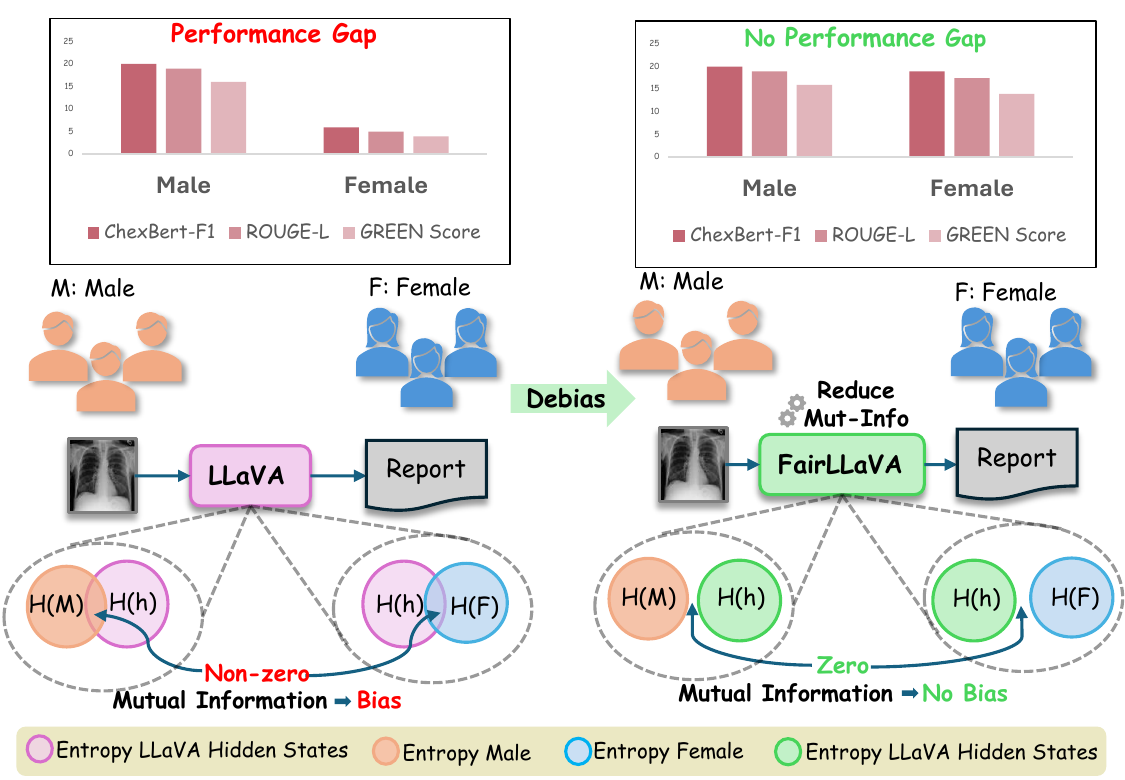}
        \caption{\textbf{FairLLaVA reduces performance disparities}. LLaVA hidden states contain demographic shortcuts (non-zero Mutual Information (MI) between hidden states and demographic attributes) that lead to lower performance for “Female". \textbf{FairLLaVA minimizes this MI promoting demographic-invariant representation learning, therefore reducing the performance gap.}}
        \label{fig:Fig1}
\end{figure}

Standard fairness techniques, such as frequency-based resampling~\cite{resample} and reweighting~\cite{reweighting, reweighting2}, aim to mitigate biased performance gaps by treating disparities as mere count imbalances. However, in MLLMs, the gaps are driven by intersectional, cross-attribute dependencies that this assumption overlooks. Ranking-based methods~\cite{nature_evaluating} require repeated inference to score and sort the entire training set, incurring substantial computational overhead.
Adversarial-classifier methods~\cite{dear_CVPR} add a separate pretrained discriminator, which, in our experiments, shows that it triggers catastrophic forgetting of clinical knowledge, evidenced by degraded overall performance (detailed in supplementary). 

Standard fairness evaluation in discriminative tasks cannot be directly applied to MLLMs either. Typically, fairness can be quantified directly on calibrated scores or binary decisions (e.g., TPR/FPR gaps, demographic parity difference, equalized odds, calibration error), which makes disparity attribution and mitigation comparatively simpler~\cite{fairmedfm, fairclip2024, fairmoe, fairtune, discrim_fair_survey_1, Tan2024FairerAI, calssification_fair_metric}. By contrast, open-ended text generation allows many valid phrasings of the same finding, and clinically important omissions can be just as harmful as incorrect mentions. Fairness, therefore, must be evaluated over the entire generated text, using semantic-level metrics and representations, rather than a single calibrated score or label.

Fairness challenges in Medical MLLMs across demographic groups have been documented, with systematic evaluations revealing performance gaps by age and additional disparities by sex and race~\cite{cares2024}. Notably, these disparities cannot be explained by sample counts alone: on MIMIC-CXR~\cite{mimic1,mimic2}, several MLLMs perform worse for patients labeled as “White” despite this being the largest cohort (detailed in the supplementary material).
{These gaps can stem from compounding factors: (i) medical images \emph{leak sensitive attributes}, models can predict self-reported race from radiology images with high AUC, even under corruptions, indicating systematically encoded demographic signal~\cite{gichoya2022race}; (ii) \emph{acquisition factors} (e.g., projection, device/site mix) correlate with demographics and mediate both predictability and downstream performance~\cite{lottter2024acq}; and (iii) \emph{imbalance} and \emph{prevalence shifts} across groups can amplify error gaps~\cite{cares2024}.}


Despite the above-mentioned concerns regarding gaps in \emph{fairness} disparities across demographic groups~\cite{cares2024, fairbench}, fairness and bias mitigation in MLLMs remain underexplored, threatening their trustworthy adoption in healthcare.  
To this point, we propose \textbf{FairLLaVA}, a fairness-aware, parameter-efficient finetuning strategy applied during visual instruction tuning. As shown in~\cref{fig:Fig1}, FairLLaVA adds a lightweight, architecture-agnostic mutual information regularizer that reduces dependence between multi-modal hidden states and sensitive attributes (age, sex, race), steering the model toward diagnostic cues instead of demographic shortcuts. FairLLaVA strikes a balance by reducing fairness gaps while preserving overall performance unlike traditional reweighting or resampling, which often improve one subgroup at the expense of others. Concretely, we jointly optimize standard instruction following with \textbf{demographic-invariant representation learning} via mutual information minimization, making the method plug-and-play with parameter-efficient finetuning (\cref{fig:methods,alg:fair-mllm}).

\noindent Our contributions are summarized as follows-
\begin{itemize}
    \item We propose FairLLaVA, a debiasing fine-tuning strategy that uses a mutual-information regularizer to remove demographic shortcuts from an MLLM’s hidden representations. Designed for models with enormous parameter counts and prohibitive fine-tuning costs, FairLLaVA is architecture-agnostic and operates with minimal intervention to the base model, incurring only modest computational overhead.
    \item We extend the Equity Scaled Metric (ES-M) to generative models, allowing seamless integration with any language-based evaluation metric, enabling a balanced assessment of both overall performance and disparities across key demographic groups.
    \item Extensive experiments on two large-scale chest radiology report–generation datasets and one dermoscopy QA dataset show that our method outperforms existing medical MLLMs (LLaVA-Rad, MedGemma, Chexagent), general-purpose MLLMs (LLaVA, Qwen, DeepSeek), and prior fairness-oriented baselines in equity scaled metrics.

\end{itemize}

%% file: sec/2_related_work.tex
\section{Related Work}
\label{fairness_definition}
Classical fairness methods can be divided into frequency-based and feature-based.
Frequency-based debiasing methods~\cite{reweighting,reweighting2,resample} attribute fairness gaps to class imbalance and mitigate them via loss re-weighting or oversampling; however, this premise does not need to hold in many settings.
Feature-based approaches seek to mitigate biases encoded within the model’s internal representations. ~\cite{dear_CVPR} addressing the fairness issue uses a pretrained adversarial classifiers to de-bias the pretrained backbone; however, it can lead to catastrophic forgetting if hyper-parameters are not tuned carefully. ~\cite{fairclip2024} regularizes image-text similarity distributions to align subgroup and overall scores. Both methods incur extra intervention via additional pretraining or higher computational cost. FairLLaVA, by contrast, handles MLLMs’ large parameter counts and expensive fine-tuning while balancing fairness gaps and overall performance.

Fairness in LLMs typically focuses on the frequency of explicit demographic phrases in generated text~\cite{llm_fair_1, llm_fair_2, toward_aistat, llm_fair_3}. ~\cite{llm_fair_1} proposes content-conditional model that for the same content (e.g. “nurse”), forces embeddings of sensitive samples (e.g., “he/she”) to be equidistant from a neutral form, promoting attribute-invariant representations.~\cite{llm_fair_2} proposes an adversarial prompt-based debiasing method that adds learnable tokens to sensitive text queries and trains an adversary over image-text similarity matrices, to preserve utility so that score distributions across demographic attributes for the same query become indistinguishable.~\cite{toward_aistat} proposes MI-based technic via importance sampling over the output distribution of LLM to reduce biases from output. However, they need extra LLM model to stabilize the debiasing signal on volatile output distribution. ~\cite{llm_fair_3} mitigates demographic bias via limited, content-aware interventions that minimize information flow from sensitive attributes by editing only bias-inducing tokens to preserve fluency. However, all~\cite{llm_fair_1, llm_fair_3, llm_fair_2, toward_aistat} define fairness at word/ lexical level which cannot be directly applied to problem setting in this work.

~\cite{poulain2024aligning} mitigates bias by creating teacher-ranked preference pairs from prompts with demographic swaps and finetuning, unlike our MI-based representation debiasing without demographic terms in the queries and does not need curating preference data.~\cite{nature_evaluating} proposes to train on the ranked responses from auxiliary RRG model to reduce the disparities, however, it incurs high computational cost to create multiple ranking pairs on the whole training data. On the contrary, FairLLaVA only adds low computational overhead and does not need additional data augmentation.

%% file: sec/4_Methods.tex
\section{Methods}
\label{sec:Methods}

\subsection{Visual Language Alignment}
Multi-modal LLMs~\cite{llava, llavarad} extend text-only backbones~\cite{vicuna2023} by encoding an image as visual tokens and combining them with text instruction tokens. A single LM backbone then produces next-token probabilities, enabling auto-regressive generation of responses grounded in the image and the given instruction.

\begin{figure*}[th!]
    \centering
    \includegraphics[width=\textwidth]{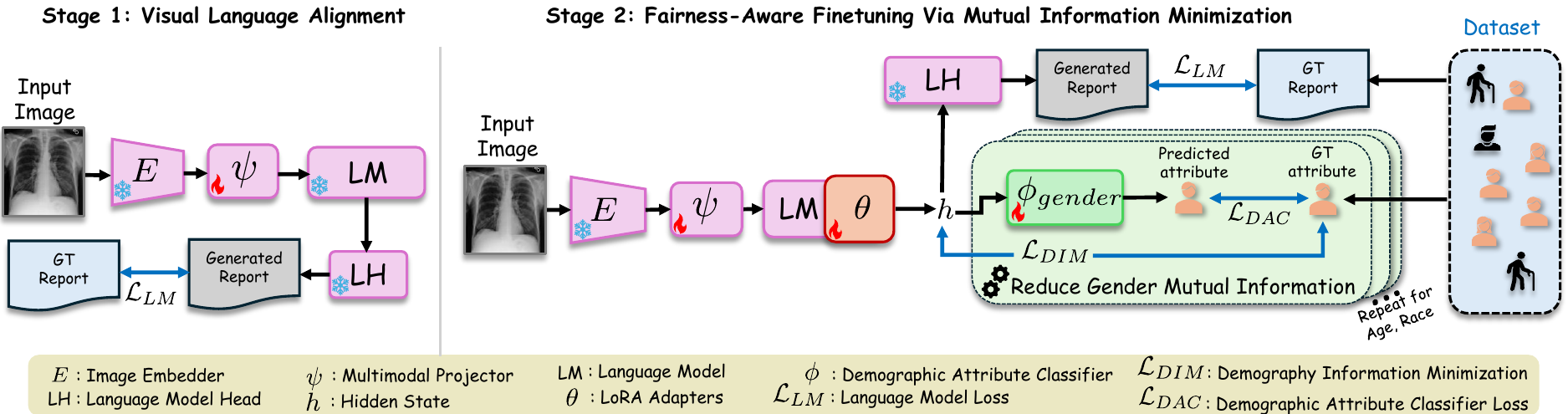}
    \caption{\textbf{FairLLaVA Overview.} \textbf{Stage 1}:  We finetune multi-modal projector $\psi$ to align the image embeddings with Language Model (LM) by optimizing standard LM CE loss $\mathcal{L}_{LM}$. Image encoder and LM are frozen.  \textbf{Stage 2}: We learn attribute-invariant representations by finetuning LoRA adapters $\theta$ on the LM’s Transformer decoder blocks while freezing the LM backbone and image encoder, pooled hidden states $h$ are fed to a mutual-information (MI) estimator with a variational demographic-attribute classifier (DAC) denoted as ${\phi}$ that predicts demography attribute from $h$. No pretrained classifier is required: the DAC is trained simultaneously with cross-entropy \(\mathcal{L}_{\text{DAC}}\) (Eq.~\eqref{eq:l_DAC}), and during this step, gradients update only \(\phi\). We then minimize MI between demographic attribute and $h$ given by \(\mathcal{L}_{\text{DIM}}\) (Eq.~\eqref{eq:club_batch}), computed between pooled states \(h\) and attributes \(\mathbf{a}\), \({\phi}\) is frozen, and only $\theta$ and $\psi$ are updated. This way DAC exposes where leakage in $h$ for predicting $\mathbf{a}$ is coming from, $\mathcal{L}_{DIM}$ suppresses this leakage making learned features demography shortcut invariant.} 
    \label{fig:methods}
\end{figure*}

Formally, consider a dataset with N triplets,
$\mathcal{D} = \{({x}_i,{a}_i,{r}_i)\}_{i=1}^{N},$ where
$x$ is an image,
$a \in \mathcal{A}$ is a corresponding demographic attribute (e.g., gender, age, race), and ${r}_i=(r_{i,1},\ldots,r_{i,T_i})$ is relevant text describing image $x$. For sample $i$, vision encoder $E_{\text{img}}$ produces features $v=E_{\text{img}}(\mathbf{x})$ which a projector ${\psi}$ maps into the Language Model (LM) space $\tilde v=\psi(v)$.
Let $u$ denote instruction tokens and define the step-$t$ context
$S_{t}=\operatorname{concat}(\tilde v,u,r_{<t})$, the language model $\theta$ maps $S_{t}$ to logits $\ell_{t}=\theta(S_{t})$ with next token distribution,
$p_\theta(\cdot \mid x,u, r_{<t})=\operatorname{softmax}(\ell_{t})$. Henceforth, we use the notation $x$ to represent encoded image tokens in place of $\tilde v$ for simplicity. The final text sequence factorizes over $T$ steps and is given by,
\begin{equation}
p_{(\theta,\psi)}(\mathbf{r}\mid x,u)
= \prod_{t=1}^{T} p_{(\theta,\psi)}(r_t \mid x,u,\mathbf{r}_{<t}).
\end{equation}
Next token prediction loss is supervised with a standard cross-entropy objective:
\begin{equation}
\mathcal{L}_{LM}(\theta, \psi)
= -\sum_{i=1}^{N}\sum_{t=1}^{T_i}\log p_{(\theta,\psi)}\!\big(r_{i,t}\,\big|\,x_i,u,r_{i,<t}\big).
\label{eq:ce_loss}
\end{equation}
As shown in ~\cref{fig:methods}, to align the pretrained image encoder embeddings with the LM, we finetune the multi-modal projector $\psi$ with supervision from $\mathcal{L}_{LM}$ in Eq.~\eqref {eq:ce_loss}. 
Since text instruction $u$ is fixed over all samples in the dataset and does not encode any demographic bias, we omit it in later sections for brevity.

\subsection{Performance Parity Metric} 
\label{subsec:es-m}
Unlike standard LLM fairness, which focuses on the presence and framing of demographic terms in outputs~\cite{llm_fair_1, llm_fair_2, llm_fair_3, toward_aistat}, we target \emph{performance parity} across groups in tasks whose outputs do not include the demographic labels explicitly ~\cite{cares2024}.
Inspired by existing fairness metrics
~\cite{fairmedfm, fairclip2024, fairmoe, fairtune, discrim_fair_survey_1, Tan2024FairerAI, calssification_fair_metric} and relevant adaptations~\cite{cares2024, nature_evaluating}, 
we redefine the performance parity metric for the text generation task. For a metric $M$ and attribute $a\in\mathcal{A}$ with group set $\mathcal{Z}_a$ (e.g. $\mathcal{Z}_{gender} = \{male, female\}$), define the per-group score:
\begin{equation}
M_{\mathbf{a}} \;=\; \mathbb{E}_{(x,r)\sim\mathcal{D}\,|\,\mathbf{a}}\!\left[\,M\!\big(\hat{r},r\big)\,\right],\ \ \mathbf{a}\in\mathcal{Z}_a.
\end{equation}
The \emph{fairness gap} is
\begin{equation}
\Delta M_a\;=\;\max_{\mathbf{a}\in \mathcal{Z}_a} M_{\mathbf{a}}\;-\;\min_{\mathbf{a}\in \mathcal{Z}_a} M_{\mathbf{a}}.
\end{equation}
The goal of the bias mitigation methods is to lower the $\Delta M_a$, narrowing the gaps across groups of demographic attribute $a$ on metric $M$. However, uniformly lower performance across groups could still yield a small $\Delta M_a$, which would misleadingly suggest fairness. Thus, overall performance must also be taken into account. Consistent with previous works~\cite{fairclip2024, toward_aistat, fairmedfm}, we observe a tradeoff between reducing fairness and overall performance (detailed in supplementary). Therefore, we extend existing equity-scaled metrics\cite{Es-AUC, fairclip2024} for the image-guided language generation tasks.
\begin{equation}
\;ES\text{-}M_a \;=\; \dfrac{M_{\mathrm{all}}}{\,1+\Delta M_a\,}\;
\end{equation}
 where, $M_{all}$ is the mean metric over all the groups of demographic attribute $a$. 
 
\subsection{Fairness-Aware Finetuning}
We hypothesize that model bias can arise from latent domain gaps across demographic groups, which may lead the model to exploit spurious correlations or shortcut patterns rather than learning true causal features. To mitigate this issue, we introduce a debiasing strategy that minimizes mutual information between learned representations and sensitive demographic attributes (e.g., gender, race, and age), thereby promoting demographic-invariant feature learning and reducing potential group disparities.

\textbf{Mutual Information.} The mutual information between the demographic attribute $\mathbf{a}$ and model hidden state $\{h_{\theta}^{l}\}$ of ${\theta}$ over the $l$-th layer is denoted as
\begin{equation}
\mathcal{I}\!\big(\mathbf{a},\,h_{\theta}^{l}(x)\,\big|\,(x, \mathbf{a})\!\sim\!\mathcal{D}\big)
= \mathbb{E}_{(x, \mathbf{a})\sim\mathcal{D}}
\left[
\log \frac{ P_{\mathcal{D}}\!\big(\mathbf{a},\,h_{\theta}^{l}(x)\big) }{ P_{\mathcal{D}}\!\big(\mathbf{a}\big)\, P_{\mathcal{D}}\!\big(h_{\theta}^{l}(x)\big)}
\right]
\label{eq:mi_eq_1}
\end{equation}
\noindent where, $P_{\mathcal{D}}\!\big(\mathbf{a},h_{\theta}^{l}(x)\big)$ is the joint distribution over $\mathbf{a}$ and $h_{\theta}^{l}(x)$, and
$P_{\mathcal{D}}\!\big(\mathbf{a}\big)$ and $P_{\mathcal{D}}\!\big(h_{\theta}^{l}\big)$ are their marginals. $h_{\theta}^{l}$ and $\mathcal{I}\!\big(\mathbf{a},\,h_{\theta}^{l}(x)\,\big|\,(x, \mathbf{a})\!\sim\!\mathcal{D}\big)$ is denoted as $h^{l}$ and $\mathcal{I}\!\big(\mathbf{a},\,h^l(x)\big)$ respectively for brevity in next sections.

\textbf{Demographic attribute classification.} Exact evaluation of $\mathcal{I}\!\big(\mathbf{a},\,h^l(x)\big)$  is intractable as the joint and marginal distributions $P_{\mathcal{D}}\!\big(\mathbf{a},h^{l}(x)\big)$ and $P_{\mathcal{D}}\!\big(h^{l}(x)\big)$ are over high-dimensional, continuous hidden states that change with model parameter ${\theta}$, additionally, we might not know the true data distribution. Hence, we minimize an upper-bound surrogate~\cite{club} $\mathcal{I}^{u}\!\big(\mathbf{a},\,h^{l}\big)$ computed via a lightweight variational demographic attribute classifier (DAC) denoted by $\phi$ mapping hidden state to demographic attribute:
\begin{equation}
\begin{split}
\mathcal{I}^{u}\!\big(\mathbf{a},\,h^{l}(x)\big)
=\;
\mathbb{E}_{(\mathbf{a},x)\sim\mathcal{D}}
\!\left[\log \phi \!\big(\mathbf{a}\,\big|\,h^{l}(x)\big) \right]
\\
-\ \mathbb{E}_{(\mathbf{a}, x),\, x'\stackrel{\text{i.i.d.}}{\sim}\mathcal{D}}
\!\left[\log \phi \!\big(\mathbf{a}\,\big|\,h^{l}(x')\big) \right].
\end{split}
\label{eq:club_upper_def}
\end{equation}

\noindent where, $x'$ is a negative sample independently drawn from mini-batch $\mathcal{B} \subset \mathcal{D}$ of size $B$ to train DAC to distinguish different demographic attributes $\mathbf{a}$:
\begin{equation}
\mathcal{L}_{DAC}(\phi)
= -\,\mathbb{E}_{(\textbf{a},x)\sim\mathcal{D}}\!\left[\log {\phi} \!\big(\mathbf{a} \mid h(x)\big)\right].
\label{eq:l_DAC}
\end{equation}
$\mathcal{L}_{DAC}$ pushes $\phi$ to learn representations helpful in identifying the demographic attribute $\textbf{a}$. 

\textbf{Demographic information minimization.} For efficiency of mutual information regularization, we pool over a set of tapped layers \(\mathcal{K}\) to get a single summarized hidden representation: \(h(x)=\frac{1}{|\mathcal{K}|}\sum_{l\in\mathcal{K}} h_l(x)\). Demographic information minimization (DIM) loss $\mathcal{L}_{DIM}$ is computed as,
\begin{equation}
\begin{split}
\mathcal{L}_{DIM}(\theta, \psi)\;=\;
\frac{1}{B}\sum_{i=1}^{B} \log {\phi} \!\big(\mathbf{a}_i \mid h(x_i)\big)
\\
\;-\;
\frac{1}{B(B-1)}\sum_{i=1}^{B}\sum_{\substack{j=1\\ j\neq i}}^{B}
\log {\phi}  \!\big(\mathbf{a}_i \mid h(x_j)\big).
\end{split}
\label{eq:club_batch}
\end{equation}
where the first term averages \emph{positive} pairs $\{(\mathbf{a}_i,x_i)\}$
and the second term averages \emph{negative} pairs
$\{(\mathbf{a}_i,x_j)\}$ with $i\neq j$, which implements the
independence between $x$ and $x'$ in Eq.\eqref{eq:club_upper_def}. This way, $\mathcal{L}_{DIM}$ updates LM backbone and MM-projector to ensure demographic-invariant representation learning, thereby reducing the shortcuts and bias against demographic attributes.

Intuitively, the mutual-information penalty treats a predictable link between the hidden state \(h(x)\) and the demographic label \(\mathbf{a}\) as leakage, pushing the representation to discard it. DAC is needed to expose where demographic information leaks into \(h(x)\) via a probe \({\phi}\) that reliably detects those cues, making the MI bound informative. Freezing \({\phi}\) and then minimizing \(\mathcal{I}^{u}(\mathbf{a},h(x))\) pushes \(h(x)\) to drop these signals, reducing shortcuts and shrinking the fairness gap \(\Delta M_{a}\).

\textbf{Parameter-efficient finetuning.} The final fairness loss constitutes $\mathcal{L}_{DAC}$ and $\mathcal{L}_{DIM}$, and language model regression loss $\mathcal{L}_{LM}$ from~\cref{eq:ce_loss} is also applied to follow the instructions and adapt the LM for the given task. The final total loss of this stage takes the form,
\begin{equation}
\label{eq:total_loss}
\mathcal{L}_{total} = \lambda_{1} \mathcal{L}_{DAC} + \lambda_{2}\mathcal{L}_{DIM} +\lambda_{3}\mathcal{L}_{LM}
\end{equation}
with $\lambda_1$, $\lambda_2$ and $\lambda_3$ controlling the fairness/utility trade-off.

\begin{algorithm}[t]
\caption{Parameter Efficient Finetuning For Fairness Via Mutual Information Minimization}
\label{alg:fair-mllm}
\DontPrintSemicolon
\KwIn{Dataset $\mathcal{D}$ with $(x,\mathbf{a},r)$ where $\mathbf{a}$ contains labels for all attributes $a\in\mathcal{A}$; fixed instruction $u$; image encoder $E_{img}$; projector params $\psi$; LM LoRA params $\theta$; tapped layers $\mathcal{K}$ of LM;  per-attribute head $\phi_a$; loss weights $\lambda_{1}, \lambda_{2}, \lambda_{3}$}

\For{\textnormal{minibatch } $\{(x_i,\mathbf{a}_i,r_i)\}_{i=1}^B \sim \mathcal{D}$}{
  \tcp{Obtain hidden states} $\{h^{l}=\theta^l(\psi(E_{img}(x_i)), u)\}_{l=1}^{|\mathcal{K}|}$ \;
  \tcp{Layer summaries}
  $h_i = \mathrm{Pool}\!\big(h^l(x_i)\big)$ \;

  \tcp{Language Model loss}
  $\mathcal{L}_{LM}(\theta, \psi) = -\sum_{i,t}\log p_{(\theta, \psi)}(r_{i,t}\mid x_i,u,r_{i,<t})$\;

  \tcp{let $\mathbf{a}_i^{(a)}$ be the label for attribute $a$ and ${\phi_a}$ its head}
  \For{$a \in \mathcal{A}$}{\

    \tcp{Demographic attribute classification}
    $\mathcal{L}_{DAC}^{(a)}(\phi) = -\frac{1}{B}\sum_i \log \phi_a\!\big(\mathbf{a}_i^{(a)} \mid \texttt{stopgrad}(h_i)\big)$\;
    
    \tcp{Demographic mutual information minimization}
    $\mathcal{L}_{DIM}^{(a)}(\theta, \psi) 
    = \frac{1}{B}\sum_i \log \phi_a\!\big(\mathbf{a}_i^{(a)} \mid h_i\big)
    \;-\; \frac{1}{B(B-1)}\sum_{i,j,\; i \neq j}\log \phi_a\!\big(\mathbf{a}_i^{(a)} \mid h_j\big)$\;
  }

  \tcp{Aggregate over attributes}
  $\mathcal{L}_{DIM} \;=\; \sum_{a\in\mathcal{A}} \mathcal{L}_{DIM}^{(a)}$ \quad
  $\mathcal{L}_{DAC} \;=\; \sum_{a\in\mathcal{A}} \mathcal{L}_{DAC}^{(a)}$\;

  \tcp{fairness-aware total loss}
  $\mathcal{L}_{total} = \lambda_{1}\mathcal{L}_{LM} + \lambda_{2}\,\mathcal{L}_{DIM} + \lambda_{3}\,\mathcal{L}_{DAC}$\;

  \tcp{Update}
  $\theta \leftarrow \theta - \eta\,\nabla_\theta \mathcal{L}_{DIM}$, \quad
  $\theta \leftarrow \theta - \eta\,\nabla_\theta \mathcal{L}_{LM}$, \;
  $\psi \leftarrow \psi - \eta\,\nabla_\psi\mathcal{L}_{DIM}$, \quad
  $\psi \leftarrow \psi - \eta\,\nabla_\psi \mathcal{L}_{LM}$, \;
  $\phi \leftarrow \phi - \eta\,\nabla_\phi \mathcal{L}_{DAC}$.\;
}
\end{algorithm}

Algorithm~\ref{alg:fair-mllm} illustrates training stage in detail. We employ parameter-efficient finetuning by adding Low-Rank Adapters (LoRA)~\cite{hu2022lora} to the transformer blocks of the language model $\theta$. Both the LM backbone and image encoder are frozen during finetuning. Projector $\psi$, LoRA adapters on $\theta$ and DAC $\phi$ are trained.
Within each iteration, we first optimize $\mathcal{L}_{DAC}$ and apply stop gradient on $h(x)$, and then freeze $\phi$ and only update LoRA $\theta$ and projector $\psi$ using $\mathcal{L}_{DIM}$ and $\mathcal{L}_{LM}$  (Algorithm~\ref{alg:fair-mllm}).

%% file: sec/5_experiments.tex
\newcommand{\quotes}[1]{``#1''}

\section{Experiments}
\label{sec:Results}

\begin{table*}[ht!]
\centering
\begingroup
\setlength{\tabcolsep}{3pt}   
\renewcommand{\arraystretch}{0.95} 

\newcommand{\rt}[1]{\rotatebox[origin=c]{45}{\makebox[1.6cm][c]{\scriptsize\bfseries #1}}}

\resizebox{0.95\linewidth}{!}{
\begin{tabular}{l*{12}{c}}
\toprule
\multirow{2}{*}{\textbf{Method}} &
\multicolumn{4}{c}{\textbf{Race}\,$\uparrow$} &
\multicolumn{4}{c}{\textbf{Age Group}\,$\uparrow$} &
\multicolumn{4}{c}{\textbf{Gender}\,$\uparrow$} \\
\cmidrule(lr){2-5}\cmidrule(lr){6-9}\cmidrule(lr){10-13}
& \rt{ES-BLEU-1} & \rt{ES-BLEU-4} & \rt{ES-RadGraph-F1} & \rt{ES-GREEN}
& \rt{ES-BLEU-1} & \rt{ES-BLEU-4} & \rt{ES-RadGraph-F1} & \rt{ES-GREEN}
& \rt{ES-BLEU-1} & \rt{ES-BLEU-4} & \rt{ES-RadGraph-F1} & \rt{ES-GREEN} \\
\midrule
\textbf{LLaVA}~\cite{llava} & 2.90 & 1.12 & 1.63 & 2.47 & 2.61 & 0.81 & 0.95 & 1.36 & 8.42 & 0.87 & 5.19 & 6.73 \\
\textbf{LLaVA-Rad}~\cite{llavarad} & 5.29 & 2.14 & \underline{4.14} & 3.97 & 8.28 & \underline{3.51} & 1.42 & 2.00 & \underline{28.06} & \textbf{12.53} & 9.24 & 10.88 \\
\textbf{MedGemma-4B}~\cite{sellergren2025medgemma} & 2.36 & 0.95 & 1.85 & \underline{5.05} & 3.17 & 1.16 & 1.51 & 3.03 & 14.18 & 1.55 & 7.27 & 7.51 \\
\textbf{MedGemma-27B}~\cite{sellergren2025medgemma} & 2.01 & 1.72 & 2.82 & 3.92 & 3.98 & 1.17 & 1.55 & \textbf{3.35} & 17.80 & 1.82 & 6.64 & 8.92 \\
\textbf{Qwen2.5-7B}~\cite{yang2025qwen3} & 7.52 & 1.47 & 2.78 & \textbf{5.59} & 6.16 & 1.37 & 1.58 & \underline{3.27} & 16.92 & 1.73 & 8.28 & 12.27 \\
\textbf{DeepSeek-VL2}~\cite{guo2025deepseek} & 4.09 & 1.47 & 2.18 & 3.27 & 3.00 & 1.23 & 1.66 & 2.51 & 11.35 & 1.66 & 6.20 & 9.37 \\
\textbf{CheXagent}~\cite{chen2024chexagent} & 4.70 & 1.05 & 1.73 & 1.49 & 2.21 & 0.73 & 0.89 & 1.37 & 9.11 & 1.04 & 4.17 & 3.46 \\
\midrule
\textbf{Reweighting-All}~\cite{reweighting} & 1.81 & 1.73 & 3.31 & 3.75 & 7.36 & 3.20 & 2.17 & 1.53 & 11.72 & 6.92 & 9.88 & \textbf{23.42} \\
\textbf{Resampling-All}~\cite{resample} & \underline{8.71} & \underline{2.32} & 2.95 & 2.13 & \underline{12.54} & 3.40 & 3.01 & 2.03 & \textbf{30.38} & \underline{11.13} & 15.97 & \underline{16.11} \\
\textbf{Adv. MLP Classifier-All}~\cite{dear_CVPR} & 3.92 & 1.08 & 2.35 & 2.34 & 6.74 & 1.11 & \underline{4.03} & 0.76 & 14.89 & 1.57 & 6.80 & 7.15 \\
\rowcolor{mygray}
\textbf{FairLLaVA-All} & \textbf{13.36} & \textbf{8.65} & \textbf{6.34} & 3.11 & \textbf{21.89} & \textbf{6.93} & \textbf{4.06} & 2.78 & 24.89 & 9.60 & \textbf{19.40} & 12.27 \\
\bottomrule
\end{tabular}}
\endgroup
\caption{\textbf{Equity-Scaled metrics on MIMIC-CXR} computed as $M_{\text{all}}/(1+\Delta M)$ for Race, Age, and Gender. (All) indicates all three demographic attributes are used for debiasing. Our joint debiasing (FairLLaVA–All) yields the state-of-the-art improvements, obtaining 7 of the 12 best ES-scores across the comprehensive semantic clinical evaluations including BLEU, RadGraph-F1, and GREEN. Best scores are in \textbf{bold} and second best \underline{underlined}.}
\label{tab:f_mimic_equity_scaled_all}
\end{table*}

\subsection{Datasets}
\label{subsec:datasets}
We evaluate FairLLaVA across multiple modalities, including grayscale chest X-ray report generation datasets with broad demographic diversity and an RGB skin-lesion dataset, demonstrating that the method generalizes beyond a single imaging modality. Dataset distributions and pre-processing details are given in the supplementary.
 
We use the \textbf{MIMIC-CXR} dataset~\cite{mimic1, mimic2} from~\cite{llavarad}, which restructures free-text radiology reports into \emph{Indication}, \emph{Findings}, and \emph{Impression} sections using GPT-4~\cite{openai2023gpt4}. It contains about 400,000 training images and 2,500 test images from 213,365 and 3,041 studies, respectively. We use the demographic attributes \emph{Age}, \emph{Race}, and \emph{Gender}. 

The original \textbf{PadChest} dataset~\cite{bustos2020padchest} contains chest X-rays with radiology reports in Spanish. We use the ChatGPT-translated English version from~\cite{stanford_rrg_challenge} and preprocess the translated reports with OpenBioLLM-70B~\cite{OpenBioLLMs-1} to extract \emph{Findings} for fine-tuning. The dataset contains about 87{,}000 training images and 8{,}000 test images. We analyze fairness gaps across the demographic attributes \emph{Age} and \emph{Gender}.

We use the \textbf{HAM10000} dataset~\cite{ham10k}, a dermatoscopic RGB skin-lesion benchmark containing 10,015 images across seven diagnostic categories. The dataset provides demographic metadata including "Age" and "Gender". Following SelfSynthX~\cite{selfsynthx}, we use the GPT-4-generated diagnostic descriptions released with their setup. We split the dataset into 8000 train and rest as test respectively.

\subsection{Implementation Details}
We use Vicuna-7b-v1.5~\cite{vicuna2023} as our base language model and BioMedCLIP~\cite{biomedclip} as the image encoder which is trained on large-scale multimodal biomedical data ($>$15M image-text pairs). For all datasets, we finetune our models for one epoch. All the models are trained on 8 NVIDIA RTX A6000 GPUs. It takes about 17 hours for training on MIMIC-CXR~\cite{mimic2} and 4 hours on the PadChest~\cite{bustos2020padchest} dataset. 
Adding the mutual information estimator (a small variational MLP) is lightweight; its cost scales with the number of classes per attribute. Even on MIMIC-CXR, which has many (14) attribute categories, this overhead remains modest ($\sim$57K parameters). Check supplementary for training stability experiments with different DAC loss contributions.

\begin{table*}[ht!]
\centering
\begingroup
\setlength{\tabcolsep}{3pt}   
\renewcommand{\arraystretch}{0.95} 

\newcommand{\rt}[1]{\rotatebox[origin=c]{45}{\makebox[1.6cm][c]{\scriptsize\bfseries #1}}}

\resizebox{0.95\linewidth}{!}{
\begin{tabular}{l*{12}{c}}
\toprule
\multirow{2}{*}{\textbf{Method}} &
\multicolumn{4}{c}{\textbf{Race}\,$\uparrow$} &
\multicolumn{4}{c}{\textbf{Age Group}\,$\uparrow$} &
\multicolumn{4}{c}{\textbf{Gender}\,$\uparrow$} \\
\cmidrule(lr){2-5}\cmidrule(lr){6-9}\cmidrule(lr){10-13}
& \rt{ES-BLEU-1} & \rt{ES-BLEU-4} & \rt{ES-RadGraph-F1} & \rt{ES-GREEN}
& \rt{ES-BLEU-1} & \rt{ES-BLEU-4} & \rt{ES-RadGraph-F1} & \rt{ES-GREEN}
& \rt{ES-BLEU-1} & \rt{ES-BLEU-4} & \rt{ES-RadGraph-F1} & \rt{ES-GREEN} \\
\midrule
\textbf{Reweighting-Race}~\cite{reweighting} & 3.38 & 1.89 & 4.13 & 3.27 & 13.15 & \underline{4.08} & \underline{2.99} & 2.09 & 15.38 & 8.28 & \underline{22.05} & 11.78 \\
\textbf{Resampling-Race}~\cite{resample} & \textbf{12.53} & \textbf{3.81} & \underline{5.00} & \underline{3.40} & \underline{14.09} & \underline{4.08} & 2.81 & \textbf{2.15} & \underline{20.62} & \underline{8.97} & 20.33 & \underline{14.26} \\
\rowcolor{mygray}
\textbf{FairLLaVA-Race} & \underline{5.40} & \underline{2.91} & \textbf{7.32} & \textbf{4.43} & \textbf{14.85} & \textbf{5.58} & \textbf{3.24} & \underline{2.10} & \textbf{25.40} & \textbf{11.15} & \textbf{27.33} & \textbf{15.05} \\
\faintmidrule
\textbf{Reweighting-Age}~\cite{reweighting} & \textbf{10.65} & \underline{3.49} & 3.24 & 2.84 & \underline{15.22} & \underline{5.26} & \textbf{3.93} & \underline{2.50} & \underline{19.73} & \textbf{11.65} & \textbf{18.33} & \textbf{14.14} \\
\textbf{Resampling-Age}~\cite{resample} & 7.43 & 2.80 & \textbf{6.26} & \underline{3.29} & 12.52 & 3.92 & 3.31 & 2.18 & \textbf{22.28} & 7.68 & 14.09 & \underline{10.91} \\
\rowcolor{mygray}
\textbf{FairLLaVA-Age} & \underline{10.29} & \textbf{15.51} & \underline{4.97} & \textbf{4.33} & \textbf{17.80} & \textbf{5.48} & \underline{3.64} & \textbf{2.59} & 17.88 & \underline{7.98} & \underline{15.05} & 10.42 \\
\faintmidrule
\textbf{Reweighting-Gender}~\cite{reweighting} & \underline{6.58} & \textbf{5.60} & \underline{3.24} & \textbf{6.02} & \underline{9.62} &\underline{3.16} & \underline{2.71} & 2.03 & \textbf{25.78} & \textbf{13.19} & \textbf{24.21} & \textbf{15.92} \\
\textbf{Resampling-Gender}~\cite{resample} & 4.45 & 1.80 & 3.12 & 4.01 & 5.58 & 2.35 & 2.04 & \underline{2.06} & 8.29 & 0.04 & 14.46 & 11.47 \\
\rowcolor{mygray}
\textbf{FairLLaVA-Gender} & \textbf{7.22} & \underline{2.70} & \textbf{8.46} & \underline{5.93} & \textbf{13.03} & \textbf{4.95} & \textbf{3.06} & \textbf{2.19} & \underline{25.43} & \underline{9.79} & \underline{23.09} & \underline{11.53} \\
\bottomrule
\end{tabular}}
\endgroup
\caption{\textbf{Equity-Scaled metrics on MIMIC-CXR when individual demographic attributes are de-biased.} Our targeted variants achieve the top ES scores on their respective attributes and show beneficial spillover to others.}
\label{tab:f_mimic_equity_scaled_individual}
\end{table*}

\subsection{Evaluation Protocol} \noindent \textbf{Baselines:}
For comprehensive evaluation and benchmarking, we include diverse baselines spanning $\sim$4–27B parameters: Qwen-2.5-7B~\cite{yang2025qwen3}, DeepSeek-VL2-7B~\cite{guo2025deepseek}, LLaVA~\cite{llava}, LLaVA-Rad-7B~\cite{llava}, MedGemma-4B~\cite{sellergren2025medgemma}, MedGemma-27B~\cite{sellergren2025medgemma}, and CheXagent-8B~\cite{chen2024chexagent}, to assess fairness-gap reduction across compact, mid-size, and large models. We also benchmark~\cite{nature_evaluating}, which, to the best of our knowledge, is the only approach explicitly designed to reduce the fairness gaps in MLLMs. Additionally, we include classical remedies for dataset bias arising from distributional skewness: resampling~\cite{resample} and reweighting~\cite{reweighting2, reweighting}. Resampling balances the training data with respect to the sensitive attribute by over-sampling underrepresented classes and under-sampling overrepresented classes, resulting in a dataset that is uniform across the demographic attribute label. Reweighting multiplies the sample training losses by their corresponding weights, which are inversely proportional to the frequency of the samples of that demographic class label in the training data. Inspired by~\cite{dear_CVPR}, we also implement an adversarial-learning baseline that uses a pretrained MLP classifier to debias the representations, denoted as “Adv. MLP Classifier” in~\cref{tab:f_mimic_equity_scaled_all}.

\noindent \textbf{Target Demographic Groups:}
As discussed in~\cref{subsec:datasets}, for the MIMIC-CXR dataset, we consider ``Age", ``Race" and ``Gender", and for PadChest and HAM10000, ``Age" and ``Gender". We first clean all datasets and remove all the records for which the demographic attributes are missing. We also exclude entries that contained information referencing a patient’s previous clinical visit. Samples are divided into three Age groups based on ranges: 0-45, 45-65, 65+, that roughly mirror standard public-health and clinical strata commonly used~\cite{CDC_DHDS_Demographics_2025}. For Gender, we consider ``Male" and ``Female" groups. For Race, there are originally 9 labels, but we only consider the 4 most occurring (``White", ``Black", ``Asian", ``Hispanic") and club the rest as ``Others". We report metrics on the baselines with the same groups, except~\cite{nature_evaluating}, for which we use demography setups evaluated in their paper.

\noindent \textbf{Metrics:}
Since different evaluation metrics focus on different aspects of the generated text~\cite{nature_evaluating}, especially in medical context~\cite{nature_evaluating, fairclip2024}, we include diverse evaluation metrics: standard Natural Language Metrics (Bleu-1~\cite{bleu}, Bleu-4~\cite{bleu}, Rouge-L~\cite{lin-2004-rouge}), clinical efficacy metrics (RadGraph-F1~\cite{jain2021radgraph}, CheXpert~\cite{irvin2019chexpert}, GREEN score~\cite{ostmeier-etal-2024-green}). BLEU-1 measures unigram (word-level) overlap with the reference, reflecting basic lexical adequacy. BLEU-4 extends this to 4-grams, making it stricter by capturing short phrases and word order, thus better reflecting fluency. ROUGE-L measures longest-sequence overlap, capturing overall sentence structure and content similarity. Additionally, for the HAM10000 dataset, we extract the predicted diagnosis from the generated report and compute classification accuracy, which reflects whether the MLLM correctly identifies the ground-truth lesion class. RadGraph-F1 evaluates clinical correctness by extracting entities (e.g., findings, anatomy) and relations (e.g., located-at, modifies) from both the generated and reference reports and computing graph overlap. CheXpert-F1 compares the presence/absence of common radiographic observations (e.g., cardiomegaly, edema) extracted by the CheXpert labeler, providing a multi-label clinical accuracy signal. GREEN is LLM based score for radiology reports that quantifies clinically-significant errors into a single metric (higher is better). RadGraph-F1 and GREEN scores are scaled into percentages. Together, these metrics balance surface-text overlap with task-specific clinical validity.

\subsection{Results}
\noindent\textbf{MIMIC-CXR.} As discussed in~\cref{subsec:es-m},
we use equity-scaled metrics to evaluate the trade-off between overall performance and performance gap across demographic attributes. \cref{tab:f_mimic_equity_scaled_all} reports results when debiasing jointly targets all attributes, whereas \cref{tab:f_mimic_equity_scaled_individual} varies which single attribute is targeted (e.g., “FairLLaVA–Age” minimizes MI only for Age). Given the intersectional relationships of subgroups \cite{nature_evaluating,fairclip2024}, we report fairness gaps for all attributes regardless of which one is used during debiasing. As shown in \cref{tab:f_mimic_equity_scaled_individual}, targeting a single attribute typically improves its own equity-scaled score (i.e., a smaller gap) but degrades the scores of the other two attributes (e.g., debiasing Age lowers the scores for Gender), indicative of a sensitive trade-off. In contrast, \cref{tab:f_mimic_equity_scaled_all} shows that jointly debiasing all attributes yields stronger, more balanced performance across Age, Gender, and Race. Across equity-scaled text and clinical metrics, our method delivers the most consistent wins, achieving 7 of 12 best ES-scores, with large margins on ES-BLEU and universal gains on ES-RadGraph-F1. Unlike classical reweighting/resampling, which often boosts one attribute at the expense of the others, joint debiasing (FairLLaVA-All) yields simultaneous gains for Race, Age, and Gender equity-scaled metrics. For GREEN, some general (non-debiasing) methods appear better for Age and Race, but this is largely due to significantly worse overall performance, coupled with smaller gaps, an artifact of the trade-off rather than genuine fairness gains. We include overall scores and raw fairness gaps in the supplementary.

Even in single-attribute debiasing (\cref{tab:f_mimic_equity_scaled_individual}), our variants remain competitive beyond the targeted attribute. For example, FairLLaVA-Race attains the best ES-BLEU-1, ES-BLEU-4 and ES-RadGraph on both Age and Gender groups; FairLLaVA–Gender yields the strongest Age ES scores across all four metrics.

\cref{tab:cmapre_nature} presents CheXpert-F1 comparisons with~\cite{nature_evaluating}, using the same setup~\cite{nature_evaluating}: ``Black" and ``White" for Race, and ``0-65" and ``65+" for Age. Notably, FairLLaVA consistently outperforms ~\cite{nature_evaluating} by a large margin on the clinically oriented CheXpert-F1 score. Since ES metric can have high variance, we verify the 95\% confidence intervals on-average remain high as compared to LLaVA-Rad. Please check supplementary, which in addition includes more settings of cross sectional and individual fairness gap analysis.

\begin{table}[h!]
\centering
\newcommand{\rt}[1]{\rotatebox[origin=c]{45}{\makebox[1.6cm][c]{\scriptsize\bfseries #1}}}

\resizebox{0.8\linewidth}{!}{
\begin{tabular}{l*{3}{c}}
\toprule
Method &
\multicolumn{1}{c}{\textbf{Race}\,$\uparrow$} &
\multicolumn{1}{c}{\textbf{Age Group}\,$\uparrow$} &
\multicolumn{1}{c}{\textbf{Gender}\,$\uparrow$} \\
\midrule
\textbf{Chen et al.}~\cite{nature_evaluating} & 24.06 & 23.85 & 24.13 \\
\rowcolor{mygray}
\textbf{FairLLaVA} & \textbf{69.21} & \textbf{68.70} & \textbf{69.38} \\
\bottomrule
\end{tabular}}
\caption{Equity-scaled CheXpert-14 F1 (higher is better) on MIMIC-CXR compared with~\cite{nature_evaluating}.}
\label{tab:cmapre_nature}
\end{table}

\begin{table*}[h!]
\centering
\setlength{\tabcolsep}{3pt}   
\renewcommand{\arraystretch}{0.95} 

\newcommand{\rt}[1]{\rotatebox[origin=c]{45}{\makebox[1.6cm][c]{\scriptsize\bfseries #1}}}

\resizebox{0.95\linewidth}{!}{
\begin{tabular}{l*{12}{c}}
\toprule
\multirow{2}{*}{\textbf{Method}} &
\multicolumn{4}{c}{\textbf{Age Group}\,$\uparrow$} &
\multicolumn{4}{c}{\textbf{Gender}\,$\uparrow$} &
\multicolumn{4}{c}{\textbf{Overall}\,$\uparrow$} \\
\cmidrule(lr){2-5}\cmidrule(lr){6-9}\cmidrule(lr){10-13}
& \textbf{\rt{ES-BLEU-1}} & \textbf{\rt{ES-BLEU-4}} & \textbf{\rt{ES-RadGraph-F1}} & \textbf{\rt{ES-GREEN}}
& \textbf{\rt{ES-BLEU-1}} & \textbf{\rt{ES-BLEU-4}} & \textbf{\rt{ES-RadGraph-F1}} & \textbf{\rt{ES-GREEN}}
& \textbf{\rt{BLEU-1}} & \textbf{\rt{BLEU-4}} & \textbf{\rt{RadGraph-F1}} & \textbf{\rt{GREEN}} \\
\midrule
\textbf{LLaVA}~\cite{llava} & 1.46 & 0.60 & 1.22 & 1.03 & 6.24 & 1.14 & 3.05 & 4.11 & 10.54 & 1.36 & 4.71 & 9.14 \\
\textbf{LLaVA-Rad}~\cite{llavarad}         & 2.31 & \underline{1.51} & \underline{3.13} & 1.22 & \underline{24.52} & 6.57 & 12.90 & \underline{6.91} & \underline{25.01} & \underline{12.02} & \underline{15.35} & \underline{39.78} \\
\textbf{MedGemma-4B}~\cite{sellergren2025medgemma}       & \underline{2.62} & 0.64 & 2.41 & \underline{2.01} & 8.99  & 1.19 & 3.87  & 5.65  & 12.31 & 1.27 & 4.37 & 11.52 \\
\textbf{MedGemma-27B}~\cite{sellergren2025medgemma}      & \textbf{2.70} & 0.86 & 2.35 & 1.79 & 9.11  & 1.61 & 4.61  & 5.21  & 13.94 & 1.79 & 5.03 & 12.67 \\
\textbf{Qwen2.5-7B}~\cite{yang2025qwen3}        & 2.28 & 1.00 & 1.56 & 1.66 & 7.73  & 1.69 & 4.12  & 4.47  & 12.45 & 1.84 & 5.11 & 11.80 \\
\textbf{DeepSeek-VL2}~\cite{guo2025deepseek}       & 2.39 & 0.97 & 1.48 & \textbf{2.02} & 8.01  & 1.77 & 3.53  & 5.83  & 11.77 & 1.91 & 4.20 & 10.44 \\
\textbf{ChexAgent}~\cite{chen2024chexagent}         & 1.44   & 0.73   & 1.12   & 0.97   & 6.74    & 0.93   & 2.79    & 3.77    & 16.74   & 2.17 & 6.12   & 21.92   \\
\midrule
\textbf{Reweighting-All}~\cite{reweighting} & 0.79 & 0.66 & 1.76 & 1.10 & 4.39  & 4.50 & 4.82  & 5.10  & 14.02 & 7.42 & 14.37 & 37.84 \\
\textbf{Resampling-All}~\cite{resample} & 2.09 & 1.31 & 2.86 & 1.14 & 14.12 & \textbf{8.60} & \underline{14.20} & 5.49  & 23.72 & 11.26 & 14.34 & 38.67 \\
\rowcolor{mygray}
\textbf{FairLLaVA-All}        & 2.53 & \textbf{1.57} & \textbf{3.52} & 1.27 & \textbf{24.86} & \underline{6.61} & \textbf{15.06} & \textbf{7.11} & \textbf{25.11} & \textbf{12.03} & \textbf{15.66} & \textbf{40.03} \\
\bottomrule
\end{tabular}}
\caption{\textbf{Equity-Scaled and Overall metrics on the PadChest dataset.} Both \quotes{Gender} and \quotes{Age} are considered in debiasing. FairLLaVA-All achieves consistently higher ES metrics across demographic attributes and also the best overall performance.}
\label{tab:f_padchest_es_age_gender_overall}
\end{table*}

\begin{table}[h!]
\centering
\setlength{\tabcolsep}{3pt}            
\renewcommand{\arraystretch}{0.95}     
\newcommand{\rt}[1]{\rotatebox[origin=c]{45}{\makebox[1.6cm][c]{\scriptsize\bfseries #1}}}

\resizebox{0.95\linewidth}{!}{
\begin{tabular}{@{}lcccccc@{}}
\toprule
\multirow{2}{*}{\textbf{Method}} &
\multicolumn{3}{c}{\textbf{Gender} $\uparrow$} &
\multicolumn{3}{c}{\textbf{Age} $\uparrow$} \\
\cmidrule(lr){2-4}\cmidrule(lr){5-7}
& \rt{ES-Acc} & \rt{ES-BLEU4} & \rt{ES-ROUGE-L}
& \rt{ES-Acc} & \rt{ES-BLEU4} & \rt{ES-ROUGE-L} \\
\midrule
LLaVA~\cite{selfsynthx} & \underline{14.12} & \underline{6.81} & \underline{14.53} & 2.02 & \underline{10.43} & \textbf{9.15} \\
 Reweighting-All~\cite{reweighting} & 11.52 & 6.25 & 12.89 & 1.69 & 2.97 & 6.61\\
 Resampling-All~\cite{resample} & 11.83 & 6.24 & 10.37 & \underline{2.19} & 6.55 & 7.03\\
\rowcolor{mygray}
FairLLaVA-All      & \textbf{19.56} & \textbf{7.06} & \textbf{16.33} & \textbf{2.63} & \textbf{12.17} & \underline{9.07} \\
\bottomrule
\end{tabular}%
}
\caption{\textbf{Equity-Scaled metrics on the HAM10000 dataset.} Both \quotes{Gender} and \quotes{Age} are considered in debiasing. FairLLaVA-All achieves consistently higher ES metrics across demographic attributes.}
\label{tab:f_ham_es}
\end{table}

\begin{table}[h!]
\centering
\setlength{\tabcolsep}{3pt}            
\renewcommand{\arraystretch}{0.95}     
\newcommand{\rt}[1]{\rotatebox[origin=c]{45}{\makebox[1.6cm][c]{\scriptsize\bfseries #1}}}

\resizebox{0.95\linewidth}{!}{
\begin{tabular}{l*{8}{c}}
\toprule
\multirow{2}{*}{\textbf{Method}} &
\multicolumn{2}{c}{\textbf{Race}\,$\downarrow$} &
\multicolumn{2}{c}{\textbf{Age Group}\,$\downarrow$} &
\multicolumn{2}{c}{\textbf{Gender}\,$\downarrow$} &
\multicolumn{2}{c}{\textbf{Overall}\,$\uparrow$} \\
\cmidrule(lr){2-3}\cmidrule(lr){4-5}\cmidrule(lr){6-7}\cmidrule(lr){8-9}
& \rt{BLEU-4} & \rt{RadGraph-F1}
& \rt{BLEU-4} & \rt{RadGraph-F1}
& \rt{BLEU-4} & \rt{RadGraph-F1}
& \rt{BLEU-4} & \rt{RadGraph-F1} \\
\midrule
\textbf{FairLLaVA-first} & 3.40 & 4.42 & 2.53 & 8.6 & 1.03 & 1.09 & 13.62 & \underline{29.71} \\
\textbf{FairLLaVA-last}  & 4.48 & \underline{3.90} & 2.4 & \textbf{3.90} & 1.34 & 0.97 & 13.19 & 29.69 \\
\textbf{FairLLaVA-mean}  & \underline{3.07} & 4.52 & \underline{2.16} & 7.32 & \underline{0.48} & \underline{0.50} & \textbf{14.84} & \textbf{29.85} \\
\rowcolor{mygray}
\textbf{FairLLaVA-mid}   & \textbf{0.61} & \textbf{3.50} & \textbf{1.01} & \underline{5.60} & \textbf{0.44} & \textbf{0.47} & \underline{14.01} & 28.52 \\
\bottomrule
\end{tabular}}
\caption{\textbf{Ablation on pooling hidden states} from FairLLaVA. \emph{FairLLaVA-mean} pools first/middle/last hidden states. The \textbf{middle layer} attains a strong balance between maintaining performance and reducing gaps across attributes.}
\label{tab:f_ablation_pool_layers}
\end{table}

\noindent
\textbf{PadChest.}
\cref{tab:f_padchest_es_age_gender_overall} compares joint debiasing methods on PadChest. FairLLaVA–All demonstrates the best balance between fairness and utility, attaining state-of-the-art overall performance on all four metrics.
Specifically, FairLLaVA–All tops 5 of 8 Equity-Scaled cells and performs best on Age (ES-BLEU-4, ES-RadGraph-F1) and Gender (ES-BLEU-1, ES-RadGraph-F1, ES-GREEN). It also attains the highest overall scores across BLEU-1/4, RadGraph-F1, and GREEN. Notably, FairLLaVA–All improves clinically grounded equity, with Gender ES-RadGraph-F1 of 15.06 and Age ES-RadGraph-F1 of 3.52. Although DeepSeek-VL2 outperforms Age ES-GREEN, and MedGemma-27B narrowly outperforms Age ES-BLEU-1, their overall performance remains one of the lowest, consistent with observations on the MIMIC-CXR dataset. Remarkably, FairLLaVA–All both mitigates demographic gaps and achieves the best overall scores. 

\noindent
\textbf{HAM10000.} \cref{tab:f_ham_es} shows that FairLLaVA-All delivers the best overall equity–utility balance on HAM10000, leading 5 of 6 Equity-Scaled metrics. It performs best across all Gender ES metrics and also achieves the top Age ES-Acc and Age ES-BLEU4, while remaining competitive on Age ES-ROUGE-L. Notably, these results verify that FairLLaVA is robust not only on grayscale radiology data but also on an RGB skin-lesion dataset, supporting claim of generalization across modalities.

\noindent
\textbf{Qualitative} results are presented in supplementary. We observe that LLaVA-Rad shows biased outputs by omitting important conditions for some subgroups. FairLLaVA, on the other hand, is able to capture key findings on both subgroups of each demographic attribute due to robust, unbiased representations learned by minimizing MI.



\subsection{Ablation}
\noindent
In MLLMs, early–mid layers handle visual grounding and reasoning, while late layers perform task-specific decoding.~\cite{middle_layer1, midle_layer2}.  ~\cref{tab:f_ablation_pool_layers}
presents an ablation on the MIMIC-CXR dataset examining which hidden layer is used in debiasing. We probe the first, middle (16th), last layers, and mean pooling over the three layers to assess which layer exhibits the greatest bias toward demographic attributes. Hidden states from the middle layer (FairLLaVA–mid) minimize fairness gaps in 5/6 cells while maintaining competitive utility. The only exception is Age RadGraph-F1, where last-layer pooling is best (3.90) but degrades other gaps and overall text quality (Overall BLEU-4 = 13.19). Mean pooling (FairLLaVA–mean) gives the best overall metrics but enlarges fairness gaps, thus we debias using middle-layer states in all of our experiments.
\vspace{-0.35cm}
\section{Conclusion}
\vspace{-0.25cm}
To mitigate demographic bias in radiology report generation, we propose FairLLaVA, a parameter-efficient fine-tuning method that suppresses demographic shortcuts in hidden states via mutual information regularization. It operates as a lightweight loss plugin with minimal intervention in the underlying language model. We quantify the fairness–utility trade-off with extensive baselines and extend standard fairness metrics to an equity-scaled variant for text generation. Across benchmarks, it narrows subgroup gaps while maintaining or even improving overall report generation quality, yielding higher equity-scaled metrics than strong competing methods such as reweighting, resampling, ranking, and adversarial-classifier baselines.

\vspace{-0.05cm}
\noindent \textbf{Limitations.}
\label{sec:limitations}
Despite minimal demographic proxies, FairLLaVA still requires demographic labels at training and evaluation to quantify fairness gaps. Many clinical datasets have incomplete or unreliable demographics, limiting applicability. 
FairLLaVA targets group-level disparities rather than individual-level/cross-sectional fairness, yet, as shown in the supplementary, it still yields reductions in both.

%% file: X_suppl.tex
\clearpage
\appendix
\setcounter{section}{0}
\setcounter{figure}{0}
\setcounter{table}{0}
\renewcommand{\thesection}{\Alph{section}}
\renewcommand{\thefigure}{S\arabic{figure}}
\renewcommand{\thetable}{S\arabic{table}}

\twocolumn[
\begin{center}
    {\LARGE \bfseries FairLLaVA: Fairness-Aware Parameter-Efficient Fine-Tuning for Large Vision-Language Assistants\par}
    \vspace{0.5em}
    {\Large Supplementary Material\par}
    \vspace{1.2em}
\end{center}
]

\section{Cross-Sectional Analysis}
In the main paper, evaluation for a given demographic attribute is performed by aggregating over all subgroups of the remaining attributes. For example, when reporting results for gender, we include samples from all age groups and all race categories. This aggregate evaluation can make the effectiveness of the attribute-specific DAC appear less direct. For instance, in Table~2 in the main paper, FairLLaVA trained to debias Age or Gender sometimes obtain higher ES scores on Race than the FairLLaVA explicitly trained for Race; similarly, the Race-debiased FairLLaVA can outperform the Gender-debiased FairLLaVA on Gender ES. Results in Table~2 in the main paper are not contradictory. Two factors explain why a model debiased for one attribute can sometimes obtain a higher ES score on another attribute than the model explicitly trained for that attribute: i) demographic attributes in MIMIC-CXR are correlated (e.g., age distributions differ across sex and race), so mitigating one source of bias can indirectly reduce disparities in another. ii) ES reflects both fairness and utility. It increases not only when subgroup disparities decrease, but also when the underlying task metric remains high. Consequently, a variant may achieve a higher ES on a non-target attribute if it preserves overall performance better, even when another variant reduces that attribute-specific gap more strongly. Thus, ES should be interpreted as a balance between gap reduction and task performance, not as a direct measure of isolated debiasing.

To isolate the effect of debiasing a specific attribute, we additionally report controlled fairness gaps by fixing the other demographic attributes. Concretely, when evaluating age-related disparities, we compare age groups only within the same race-gender subgroup slice, rather than mixing samples across different races or genders. This reduces confounding from correlated demographics and provides a cleaner view of whether the method is truly reducing bias for the intended attribute. We call this cross-sectional analysis. As shown in ~\cref{tab:cross_section_gap_ablate} upper rows, under this analysis, the gap decreases the most when the corresponding attribute is explicitly debiased, confirming the intended effect. We also compare the FairLLaVA-All variant with the strongest baseline of LLaVA-Rad.

\begin{table}[t]
\centering
\scriptsize
\setlength{\tabcolsep}{3pt}
\renewcommand{\arraystretch}{0.9}

\resizebox{0.99\linewidth}{!}{%
\begin{tabular}{lcccccc}
\toprule
\multirow{2}{*}{\textbf{Method}} &
\multicolumn{2}{c}{\textbf{Race}\,$\downarrow$} &
\multicolumn{2}{c}{\textbf{Age}\,$\downarrow$} &
\multicolumn{2}{c}{\textbf{Gender}\,$\downarrow$} \\
\cmidrule(lr){2-3}\cmidrule(lr){4-5}\cmidrule(lr){6-7}
& \textbf{RG-F1} & \textbf{GREEN}
& \textbf{RG-F1} & \textbf{GREEN}
& \textbf{RG-F1} & \textbf{GREEN} \\
\midrule
FairLLaVA-Race   & 2.98 & 4.59 & 9.06 & 16.88 & 1.41 & 1.84 \\
FairLLaVA-Age    & 4.12 & 6.37 & 8.17 & 14.88 & 1.61 & 3.03 \\
FairLLaVA-Gender & 2.89 & 6.01 & 9.29 & 16.04 & 1.18 & 1.81 \\
\midrule
LLaVA-Rad          & 3.59 & 6.30 & 10.38 & 17.28 & 1.17 & 3.54 \\
\rowcolor{mygray}
FairLLaVA-All    & \textbf{3.26} & \textbf{4.78} & \textbf{7.33} & \textbf{12.89} & \textbf{1.08} & \textbf{2.74} \\
\bottomrule
\end{tabular}%
}
\caption{\textbf{Cross-sectional fairness analysis.} RG-F1 is an abbreviation of RadGraph-F1. To isolate the effect of debiasing each demographic attribute, subgroup gaps are computed while holding the remaining attributes fixed (e.g., comparing age groups within the same race--gender slice). The targeted-attribute variants cause most reduction in the gap as compared to other-attribute variants, as intended. FairLLaVA-All also holds strong under this analysis as compared to the strong baseline of LLaVA-Rad.}
\label{tab:cross_section_gap_ablate}
\end{table}

\section{Individual Performance and Counter-Factual Fairness Gaps}
To assess spurious demographic reliance beyond aggregate subgroup metrics, we perform a counterfactual fairness analysis at the individual level. The goal is to measure the fairness gap when the protected attribute varies while the underlying clinical evidence is kept as similar as possible. Concretely, for each sample, we retrieve its nearest match from a different demographic subgroup in the latent feature space, subject to two constraints: (i) the pair must share the same CheXpert label set, so that the clinical findings are matched, and (ii) the latent similarity must exceed a threshold of $0.7$, ensuring that the paired samples are visually and semantically close. For example, a female study with pleural effusion is matched to the nearest male study with the same CheXpert label i.e. pleural effusion. We then compute the fairness gap across such matched pairs. Since the paired samples are aligned in clinical content, a large gap indicates that the model is relying on demographic cues beyond the disease evidence, whereas a smaller gap suggests reduced spurious dependence on the protected attribute. We used a BiomedCLIP-CXR~\cite{biomedclip} as feature extractor. As seen in~\cref{tab:counterfactual_gaps}, FairLLaVA consistently lowers these counterfactual gaps relative to LLaVA-Rad across Race, Age, and Gender, indicating improved robustness to demographic variation under matched clinical evidence. This analysis complements population-level ES metrics by providing an individual-level fairness signal, and can also serve as an initial signal to decide which demographic attributes to de-bias and choose $\lambda$ weights.

\begin{table}[t]
\centering
\scriptsize
\setlength{\tabcolsep}{3pt}
\renewcommand{\arraystretch}{0.9}
\resizebox{0.99\linewidth}{!}{%
\begin{tabular}{lcccccc}
\toprule
\multirow{2}{*}{\textbf{Method}} &
\multicolumn{2}{c}{\textbf{Race}\,$\downarrow$} &
\multicolumn{2}{c}{\textbf{Age}\,$\downarrow$} &
\multicolumn{2}{c}{\textbf{Gender}\,$\downarrow$} \\
\cmidrule(lr){2-3}\cmidrule(lr){4-5}\cmidrule(lr){6-7}
& \textbf{RG-F1} & \textbf{GREEN}
& \textbf{RG-F1} & \textbf{GREEN}
& \textbf{RG-F1} & \textbf{GREEN} \\
\midrule
LLaVA-Rad & 13.71 & 24.45 & 27.44 & 21.44 & 17.93 & 21.60 \\
FairLLaVA &  6.65 & 23.08 & 20.34 & 17.65 & 16.92 & 18.90 \\
\bottomrule
\end{tabular}%
}
\caption{\textbf{Counterfactual fairness gaps.} FairLLaVA also reduces the individual counterfactual fairness gaps on MIMIC-CXR dataset.}
\label{tab:counterfactual_gaps}
\end{table}

\section{Hyper-parameters}
\begin{figure*}[th!]
    \centering
    \includegraphics[width=0.99\textwidth]{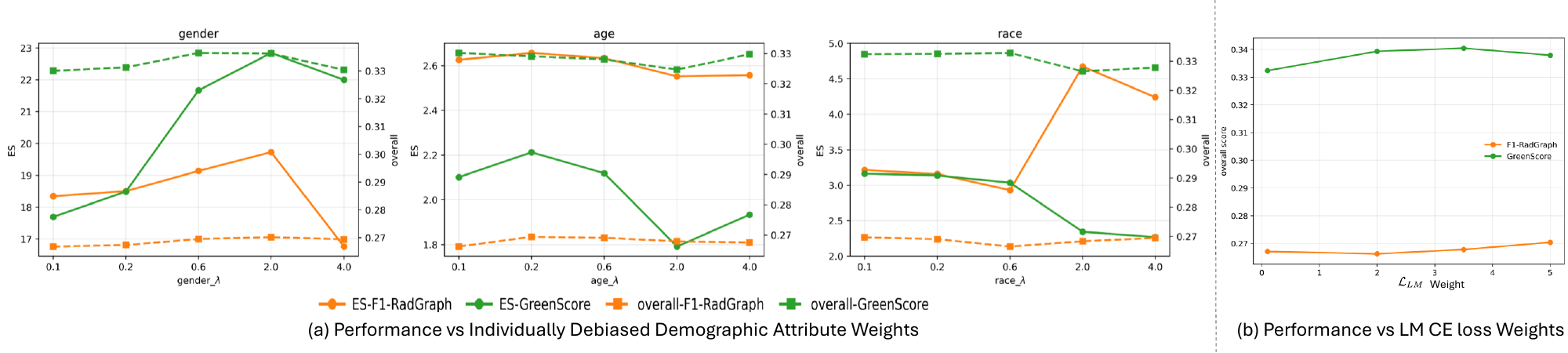}
    \caption{\textbf{Hyper Parameters Sensitivity} (a) Varying the contribution of each attribute-specific MI term to the total loss on MIMIC-CXR leads to only minor changes, indicating stable overall performance across attributes. (b) Varying the contribution of language model loss $\mathcal{L}_{LM}$ leads to minor changes in overall performance.} 
    \label{fig:train_stability}
\end{figure*}

\noindent
\paragraph{Sensitivity}
Total loss is given by~eq.10 which has many hyper-parameters. Here we study their sensitivity to the fairness gap. 
We train the DAC using a class-frequency-weighted cross-entropy loss to mitigate class imbalance, and use $\lambda$ to weight the DAC-based MI minimization term.~\cref{fig:train_stability}~(a) shows that increasing the three DAC $\lambda$ values results in only marginal changes in the equity-scaled metric (ES-M) on MIMIC-CXR. We further examine the $\lambda$ for the LM loss in~\cref{fig:train_stability}~(b). In both cases, the overall performance remains largely stable, indicating that the added components do not introduce significant sensitivity.

\noindent
\paragraph{Hyper-parameter Search}
In~\cref{tab:lambda_grid_search_mimic}, we vary $(\lambda_{\text{race}}, \lambda_{\text{age}}, \lambda_{\text{gender}})$ by increasing one attribute weight at a time to study its effect on MIMIC-CXR. While all three attributes are debiased jointly, the largest ES gain for an attribute is achieved when its corresponding $\lambda$ is assigned the highest value. In the main manuscript, we use $(0.2, 0.6, 0.1)$ to approximately reflect the relative LLaVA-Rad baseline gap ratios $(\Delta_{\text{race}} : \Delta_{\text{age}} : \Delta_{\text{gender}})$. Similarly, we set the values for PadChest as (0.0, 0.3, 0.2) and for HAM10000 as (0.0, 0.6, 0.2). 

\begin{table}[t]
\centering
\scriptsize
\setlength{\tabcolsep}{3pt}
\renewcommand{\arraystretch}{0.99}
\resizebox{0.48\textwidth}{!}{%
\begin{tabular}{lcccccc}
\toprule
\multirow{2}{*}{\textbf{$(\lambda_r,\lambda_a,\lambda_g)$}} &
\multicolumn{2}{c}{\textbf{Race}\,$\uparrow$} &
\multicolumn{2}{c}{\textbf{Age Group}\,$\uparrow$} &
\multicolumn{2}{c}{\textbf{Gender}\,$\uparrow$} \\
\cmidrule(lr){2-3}\cmidrule(lr){4-5}\cmidrule(lr){6-7}
& \textbf{RG-F1} & \textbf{GREEN}
& \textbf{RG-F1} & \textbf{GREEN}
& \textbf{RG-F1} & \textbf{GREEN} \\
\midrule
$(0.2, 0.6, 0.2)$ & 2.77 & 3.22 & 3.85 & 3.34 & 18.98 & 12.92 \\
$(0.6, 0.2, 0.2)$ & 5.62 & 3.60 & 3.42 & 1.83 & 22.45 & 12.23 \\
$(0.2, 0.2, 0.6)$ & 5.30 & 3.18 & 3.24 & 1.85 & 24.74 & 13.77 \\
\bottomrule
\end{tabular}%
}
\caption{\textbf{Effect of varying attribute-specific MI weights.} $(\lambda_r,\lambda_a,\lambda_g)$ on equity-scaled performance.}
\label{tab:lambda_grid_search_mimic}
\end{table}

\section{Handling Missing Labels}
Requiring demographic labels could be a limiting factor as we discussed in Sec. Limitations in the main paper. in this section, we show that this limitation can be effectively addressed. Recall, FairLLaVA does not need labels at inference time. For training, missing attributes can be predicted reliably on zero-shot radiology datasets: as shown in~\cref{tab:missing_demographic}, TorchXRayVision~\cite{torch-xray-vision} demographic predictors, trained on CheXpert~\cite{irvin2019chexpert} and NIH ChestX-ray14~\cite{majkowska2020chest}, achieve high AUC and low age MAE (in years) in our evaluation on both MIMIC-CXR and PadChest datasets.

\begin{table}[t]
\centering
\scriptsize
\setlength{\tabcolsep}{3pt}
\renewcommand{\arraystretch}{0.95}

\newcommand{\rt}[1]{\rotatebox[origin=c]{45}{\makebox[1.25cm][c]{\scriptsize\bfseries #1}}}

\resizebox{0.99\linewidth}{!}{%
\begin{tabular}{lcccc}
\toprule
\multirow{2}{*}{\textbf{Dataset}} &
\textbf{Sex$\uparrow$} &
\multicolumn{2}{c}{\textbf{Age$\uparrow$}} &
\textbf{Race$\uparrow$} \\
\cmidrule(lr){2-2}\cmidrule(lr){3-4}\cmidrule(lr){5-5}
& \rt{AUC} & \rt{MAE} & \rt{AgeGrp Acc} & \rt{AUC} \\
\midrule
MIMIC-CXR & 0.9549 & 6.9715 & 0.7265 & 0.8901 \\
PadChest  & 0.9841 & 6.2339 & 0.8023 & -- \\
\bottomrule
\end{tabular}%
}
\caption{\textbf{Missing demographic attribute prediction} on out of domain radiology datasets using TorchXRayVision~\cite{torch-xray-vision}.}
\label{tab:missing_demographic}
\end{table}

\section{Variance in Equity Scaled Metric}
\begin{figure*}[th!]
    \centering
    \includegraphics[width=0.99\textwidth]{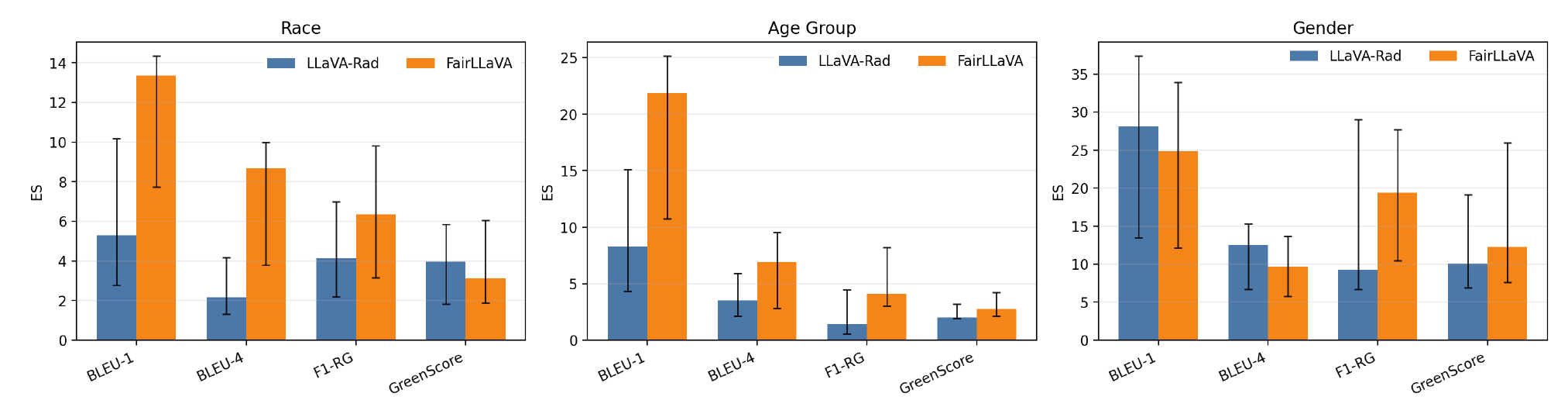}
    \caption{\textbf{95\% Confidence Intervals} of ES metric on MIMIC-CXR with bootstrap resampling ($n{=}1000$)} 
    \label{fig:ES-CI}
\end{figure*}

\Cref{fig:ES-CI} reports 95\% confidence intervals obtained via bootstrap resampling ($n{=}1000$) on MIMIC-CXR dataset. We observe that ES scores, as well as the underlying subgroup gaps, can exhibit large variance. This is expected because i) fairness gaps are calculated as difference between maximum and minimum subgroup performance, which are extreme values on both ends that inherently have high variance ii) fairness metrics are computed over smaller demographic subgroups, where class imbalance and limited sample counts can amplify estimation noise, iii) moreover, ES depends jointly on both the subgroup gap and the overall task performance, so uncertainty in either term amplifies into the ES score. For this reason, ES should be interpreted together with its confidence interval. All the quantitative results in this work, therefore, report median values. Despite this variability, on average, FairLLaVA (\cref{fig:ES-CI}) shows consistently stronger performance across most demographic attributes.

\section{Prevalence Trends}
To further assess whether FairLLaVA suppresses clinically meaningful subgroup-specific disease patterns, we compare disease prevalence in ground-truth and model-generated reports on the test set. We obtain disease labels by applying the CheXpert labeler to both the reference reports and FairLLaVA-generated reports from the test split. For each CheXpert finding and each demographic subgroup (gender, race-major, and age-group), prevalence is computed as the fraction of samples labeled positive. We then measure the prevalence shift as,
\begin{equation}
\Delta = p_{\mathrm{pred}} - p_{\mathrm{ref}},
\end{equation}
where \(p_{\mathrm{ref}}\) and \(p_{\mathrm{pred}}\) denote the reference and generated prevalence, respectively.

To focus on clinically meaningful and statistically reliable patterns, we retain only subgroup-finding pairs with subgroup size \(N > 50\) and reference prevalence \(p_{\mathrm{ref}} \geq 0.7\). We then rank these pairs by descending reference prevalence in ~\cref{tab:prevalence_preservation_filtered}. As shown in~\cref{tab:prevalence_preservation_filtered}, FairLLaVA largely preserves these strong prevalence patterns, indicating that debiasing does not simply erase important population-level disease signals.

\begin{table*}[t]
\centering
\scriptsize
\setlength{\tabcolsep}{5pt}
\renewcommand{\arraystretch}{0.95}
\resizebox{0.8\linewidth}{!}{
\begin{tabular}{llrccc}
\toprule
\textbf{Group} & \textbf{Finding} & \textbf{N} & \textbf{$p_{\mathrm{ref}}$} & \textbf{$p_{\mathrm{pred}}$} & \textbf{$\Delta$} \\
\midrule
\multicolumn{6}{l}{\textbf{Age Group}} \\
\midrule
0-44  & Pleural Effusion & 87   & 0.851 & 0.874 &  0.023 \\
44-65 & Pleural Effusion & 899  & 0.790 & 0.810 &  0.020 \\
65+    & Pleural Effusion & 1014 & 0.764 & 0.824 &  0.060 \\
0-44  & Pneumothorax     & 87   & 0.736 & 0.736 &  0.000 \\
\midrule
\multicolumn{6}{l}{\textbf{Gender}} \\
\midrule
F & Pleural Effusion & 896  & 0.823 & 0.839 &  0.017 \\
M & Pleural Effusion & 1104 & 0.745 & 0.804 &  0.060 \\
\midrule
\multicolumn{6}{l}{\textbf{Race}} \\
\midrule
Hispanic or Latino        & Pleural Effusion & 51   & 0.863 & 0.882 &  0.020 \\
Black or African American & Pleural Effusion & 427  & 0.824 & 0.874 &  0.049 \\
Hispanic or Latino        & Pneumothorax     & 51   & 0.784 & 0.804 &  0.020 \\
Asian                     & Support Devices  & 77   & 0.779 & 0.701 & -0.078 \\
White                     & Pleural Effusion & 1381 & 0.775 & 0.812 &  0.038 \\
Asian                     & Pleural Effusion & 77   & 0.701 & 0.701 &  0.000 \\
\bottomrule
\end{tabular}}
\caption{\textbf{Prevalence preservation under FairLLaVA} for subgroup-finding pairs with reference prevalence \(p_{\mathrm{ref}}\geq 0.7\) and subgroup size \(N>50\). We report the reference prevalence \(p_{\mathrm{ref}}\), generated prevalence \(p_{\mathrm{pred}}\), and prevalence shift \(\Delta = p_{\mathrm{pred}} - p_{\mathrm{ref}}\). High-prevalence findings are largely preserved across demographic groups, with generally small shifts in prevalence.}
\label{tab:prevalence_preservation_filtered}
\end{table*}

\section{Subgroup Size–Performance Correlation}
\begin{figure*}[th!]
    \centering
    \includegraphics[width=0.83\textwidth]{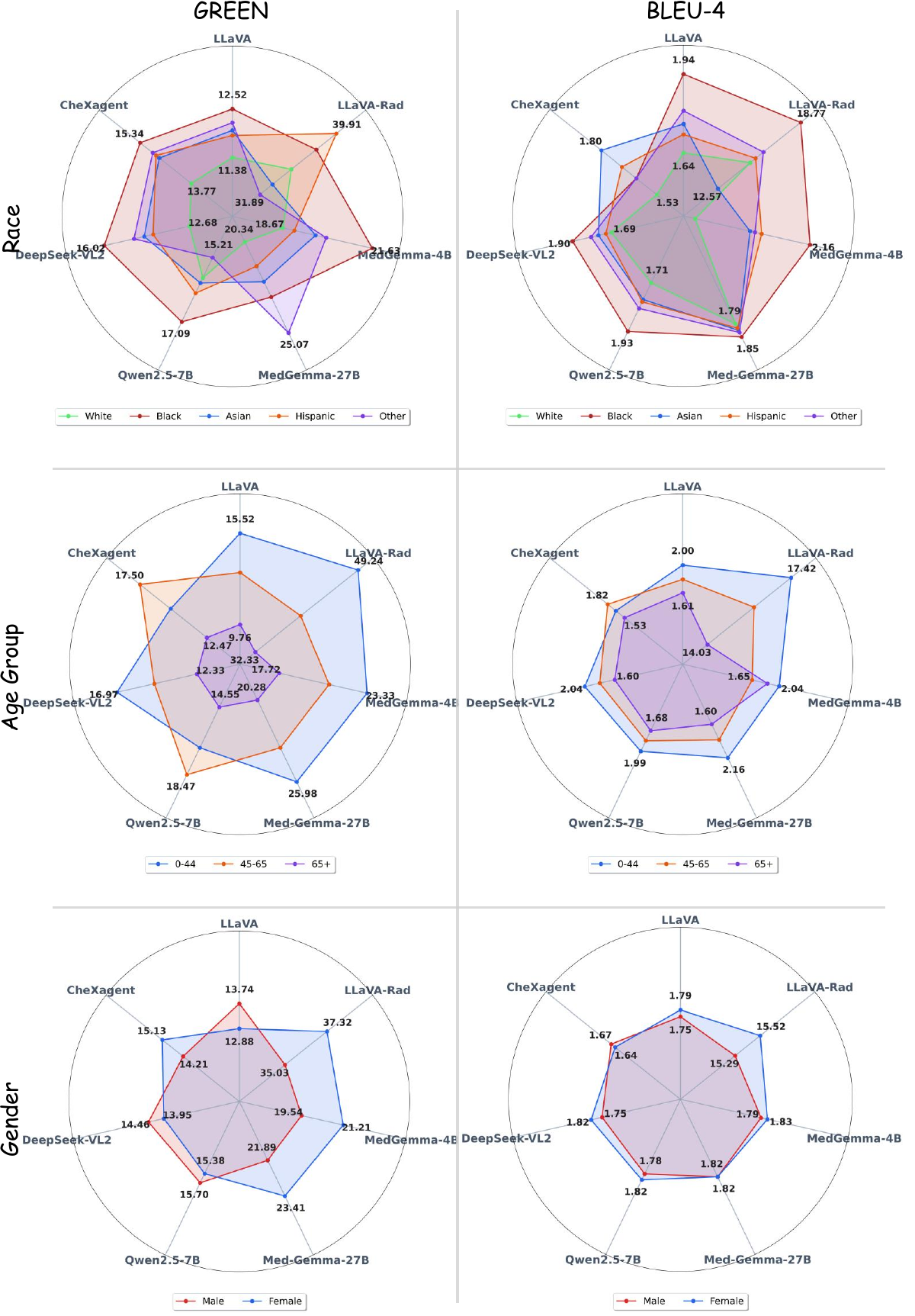}
    \caption{Sub-Group performance of baselines across “Race”, “Age” and “Gender” subgroups on MIMIC-CXR dataset on GREEN and BLEU-4 metric. We observe that the high number of counts in the train dataset does not correlate with the increased performance. Please also check~\cref{fig:dem_dist}} 
    \label{fig:mimic_subgrp}
\end{figure*}

\begin{figure*}[th!]
    \centering
    \includegraphics[width=0.83\textwidth]{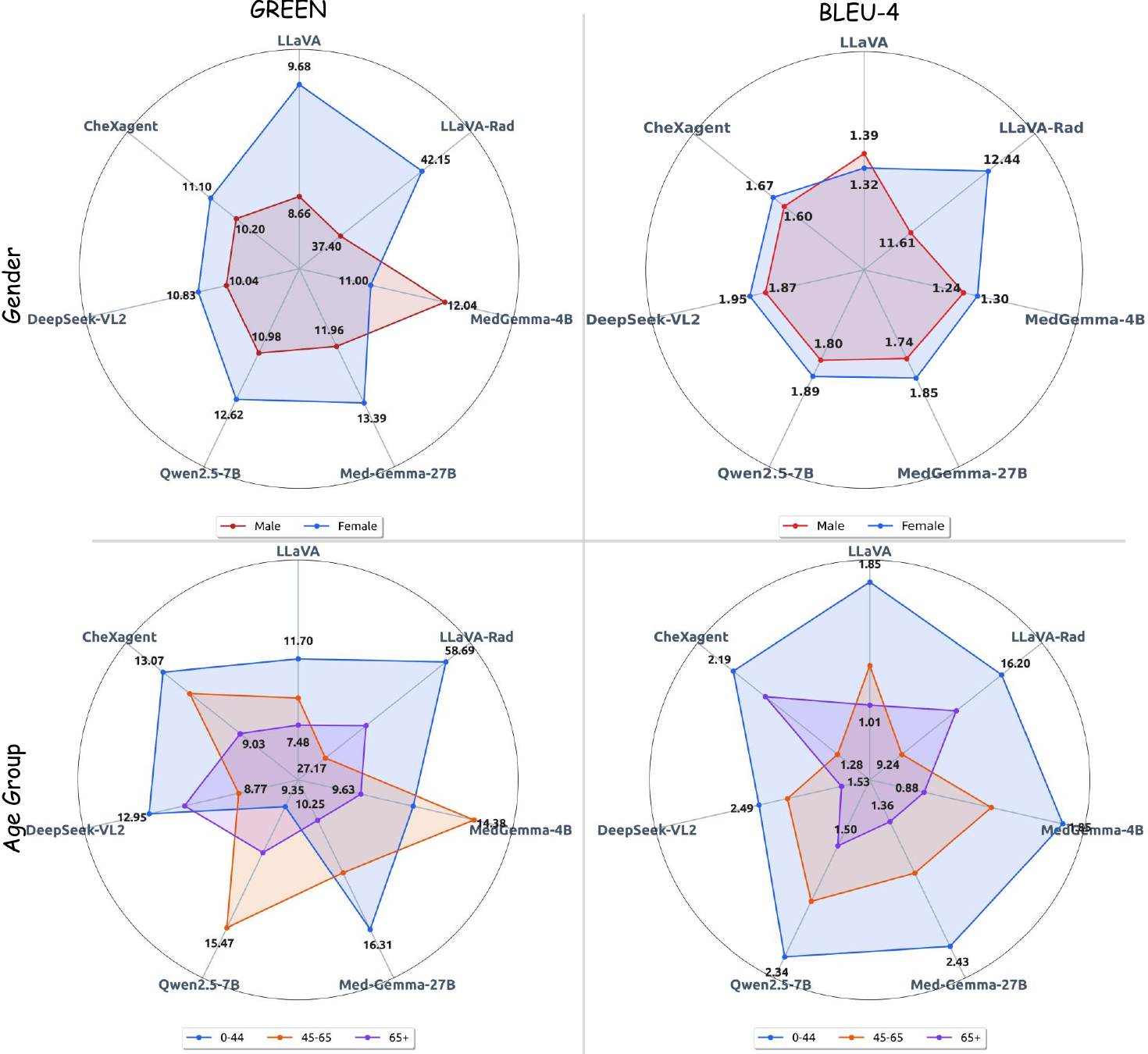}
    \caption{Sub-Group performance of baselines across “Age” and “Gender” subgroups on PadChest dataset on GREEN and BLEU-4 metric. We observe that the high number of counts in the train dataset does not correlate with the increased performance. Please also check~\cref{fig:dem_dist}} 
    \label{fig:padchest_subgrp}
\end{figure*}

\begin{figure*}[th!]
    \centering
    \includegraphics[width=\textwidth]{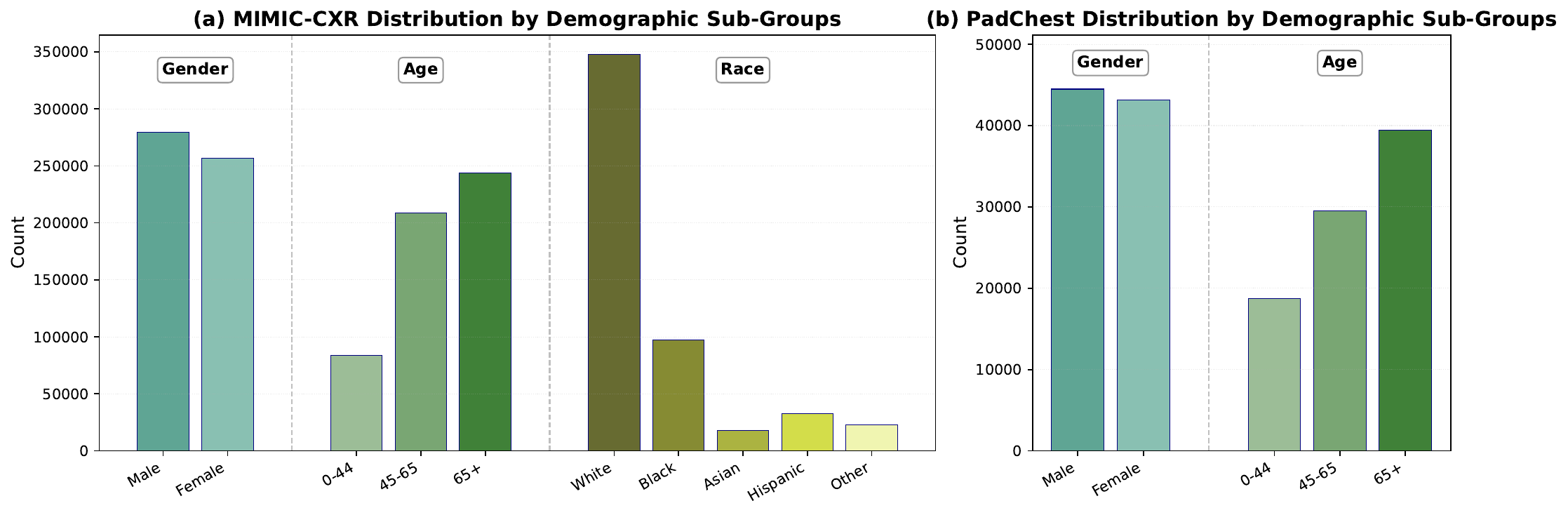}
    \caption{Distribution of counts of demographic subgroups in MIMIC-CXR and PadChest dataset train splits. Some demographic group counts are highly imbalanced, (a) MIMIC-CXR and (b) PadChest. F1-RG is an abbreviation of RadGraph-F1.} 
    \label{fig:dem_dist}
\end{figure*}

In this section we present metrics for each subgroup of the demographic attributes “Age”, “Race” and “Gender” in MIMIC-CXR and “Age”, “Gender” in PadChest in~\cref{fig:mimic_subgrp} and~\cref{fig:padchest_subgrp} respectively. We detail distribution of the samples across these subgroups in~\cref{fig:dem_dist} (a) for MIMIC CXR and~\cref{fig:dem_dist} (b) for PadChest. Datasets are quite unbalanced for some demographic attributes, especially for “Age” and “Race”. However, as pointed out in the main paper, lower sample count does not automatically mean lower performance, indicating naive classical frequency based methods might not work. For example “White” is the most frequent race in MIMIC-CXR dataset, however, it does not perform the best on any of the baselines as seen in~\cref{fig:mimic_subgrp}. But performance for the “Black” race is almost always the best, despite it being more than double less represented in the dataset. Similar results are seen on the PadChest dataset in~\cref{fig:padchest_subgrp}, for example in “Age” 65+ never gets highest performance on any baseline despite having significantly larger count of samples (on neither clinically oriented GREEN score nor classical BLEU-4 score). 

\section{Fairness Gaps and Overall Performance}

We report Equity Scaled Metric (ES-M) in the main paper as balance between the gaps between demographic groups as well as the absolute quality of the generated reports. 

In \cref{tab:fair_gaps_mimic}, we show both fairness gaps and overall performance on the MIMIC-CXR dataset.
For more general-purpose as well as medical MLLMs such as MedGemma-4B/27B~\cite{sellergren2025medgemma}, Qwen2.5-7B~\cite{yang2025qwen3}, DeepSeek-VL2~\cite{guo2025deepseek}, LaVA-Rad~\cite{llavarad}, the fairness gaps are small, but this comes with substantially lower overall performance compared to the top performing models. For example, Qwen2.5-7B attains the lowest GREEN gaps for race and age and the second-lowest for gender, yet its overall GREEN score (16.10) is less than half of FairLLaVA’s 34.32. This illustrates why gap-only metrics can be misleading: a model that is uniformly weak across all groups can appear “fair” while offering limited clinical utility, which motivates our use of ES-Metrics as a more comprehensive evaluation. 
Among classical fairness methods,  reweighting and oversampling we see distinct tradeoffs, they reduce some gaps but either leave others relatively large or noticeably degrade overall performance. The adversarial style fairness solution~\cite{dear_CVPR} has significantly lower performance especially in clinically important metrics such as overall GREEN score that drops to 9.36, nearly a four-fold decrease relative to FairLLaVA (34.32), consistent with catastrophic forgetting of clinically meaningful information. In contrast, FairLLaVA substantially lowers disparities yet keeping comparable overall performance to the best performing LLaVA-Rad, yielding a more balanced fairness–utility trade-off (demonstrated in ES-Metrics tables in the main paper).

Similarly, \cref{tab:fair_gaps_padchest} on PadChest dataset shows FairLLaVA maintains overall performance that is comparable to, and in some cases exceeds, the best-performing LLaVA-Rad model, while substantially reducing fairness gaps. It clearly outperforms other fairness approaches in terms of demographic fairness across Age and Gender, as well as overall evaluation metrics.

\begin{table*}[h!]
\centering
\begingroup
\setlength{\tabcolsep}{3pt}
\renewcommand{\arraystretch}{0.95} 

\newcommand{\rt}[1]{\rotatebox[origin=c]{45}{\makebox[1.6cm][c]{\scriptsize\bfseries #1}}}

\resizebox{0.99\linewidth}{!}{
\begin{tabular}{l*{16}{c}}
\toprule
\multirow{2}{*}{\textbf{Method}} &
\multicolumn{4}{c}{\textbf{Race}\,$\downarrow$} &
\multicolumn{4}{c}{\textbf{Age Group}\,$\downarrow$} &
\multicolumn{4}{c}{\textbf{Gender}\,$\downarrow$} &
\multicolumn{4}{c}{\textbf{Overall}\,$\uparrow$} \\
\cmidrule(lr){2-5}\cmidrule(lr){6-9}\cmidrule(lr){10-13}\cmidrule(lr){14-17}
& \rt{BLEU-1} & \rt{BLEU-4} & \rt{RadGraph-F1} & \rt{GREEN}
& \rt{BLEU-1} & \rt{BLEU-4} & \rt{RadGraph-F1} & \rt{GREEN}
& \rt{BLEU-1} & \rt{BLEU-4} & \rt{RadGraph-F1} & \rt{GREEN}
& \rt{BLEU-1} & \rt{BLEU-4} & \rt{RadGraph-F1} & \rt{GREEN} \\
\midrule

\textbf{LLaVA-Rad}
& 6.22 & 6.20 & 6.20 & 8.02
& 3.61 & 3.39 & 19.97 & 16.91
& 0.36 & 0.23 & 2.23 & 2.29
& 38.17 & 15.40 & 29.80 & 35.82 \\

\textbf{MedGemma-4B}
& 6.21 & 0.69 & 4.70 & 2.96
& 4.37 & 0.39 & 5.98 & 5.61
& 0.20 & 0.04 & 0.45 & 1.67
& 17.02 & 1.61 & 10.54 & 20.02 \\

\textbf{MedGemma-27B}
& 8.03 & 0.06 & 4.13 & 4.73
& 3.56 & 0.56 & 8.33 & 5.70
& 0.02 & 0.00 & 1.18 & 1.51
& 18.15 & 1.82 & 14.46 & 22.45 \\

\textbf{Qwen2.5-7B}
& 1.43 & 0.22 & 2.66 & 1.88
& 1.97 & 0.31 & 5.44 & 3.92
& 0.08 & 0.04 & 0.23 & 0.32
& 18.29 & 1.80 & 10.18 & 16.098 \\

\textbf{DeepSeek-VL2}
& 2.17 & 0.21 & 2.73 & 3.34
& 3.32 & 0.44 & 3.90 & 4.64
& 0.14 & 0.07 & 0.31 & 0.51
& 12.93 & 1.78 & 8.12 & 14.16 \\


\midrule

\textbf{Reweighting-All}
& 11.87 & 3.44 & 5.27 & 5.50
& 2.17 & 1.40 & 8.58 & 14.93
& 0.99 & 0.11 & 1.10 & 0.04
& 23.30 & 7.69 & 20.76 & 24.38 \\

\textbf{Resampling-All}
& 3.22 & 4.90 & 7.66 & 15.25
& 1.93 & 3.03 & 7.49 & 16.05
& 0.21 & 0.23 & 0.60 & 1.15
& 36.75 & 13.69 & 25.55 & 34.61 \\

\textbf{Adv. MLP Classifier-All}
& 4.37 & 0.74 & 3.51 & 3.00
& 2.12 & 0.70 & 1.63 & 11.32
& 0.41 & 0.20 & 0.56 & 0.31
& 21.04 & 1.88 & 10.61 & 9.36 \\

\rowcolor{mygray}
\textbf{FairLLaVA-All}
& 1.61 & 0.61 & 3.50 & 9.97
& 0.59 & 1.01 & 5.60 & 12.29
& 0.40 & 0.45 & 0.47 & 1.80
& 34.85 & 13.92 & 28.52 & 34.32 \\
\bottomrule
\end{tabular}}
\endgroup
\caption{Fairness Gaps (First three main columns across Race, Age, Gender) and Overall performance (last column) on MIMIC CXR dataset. Highlights tradeoff between Overall-Performance and Fairness-Gaps. Fairness gaps lower the better, Overall performance higher the better.}
\label{tab:fair_gaps_mimic}
\end{table*}

\begin{table*}[h!]
\centering
\resizebox{0.99\linewidth}{!}{
\begin{tabular}{l*{12}{c}}
\toprule
\multirow{2}{*}{\textbf{Method}} &
\multicolumn{4}{c}{\textbf{Age Group}\,$\downarrow$} &
\multicolumn{4}{c}{\textbf{Gender}\,$\downarrow$} &
\multicolumn{4}{c}{\textbf{Overall}\,$\uparrow$} \\
\cmidrule(lr){2-5}\cmidrule(lr){6-9}\cmidrule(lr){10-13}
& \textbf{BLEU-1} & \textbf{BLEU-4} & \textbf{RadGraph-F1} & \textbf{GREEN}
& \textbf{BLEU-1} & \textbf{BLEU-4} & \textbf{RadGraph-F1} & \textbf{GREEN}
& \textbf{BLEU-1} & \textbf{BLEU-4} & \textbf{RadGraph-F1} & \textbf{GREEN} \\
\midrule
\textbf{LLaVA-Rad}
& 9.81 & 6.96 & 3.91 & 31.52
& 0.02 & 0.83 & 0.19 & 4.76
& 25.01 & 12.02 & 15.35 & 39.78 \\
\textbf{MedGemma-4B}
& 3.70 & 0.97 & 0.81 & 4.73
& 0.37 & 0.07 & 0.13 & 1.04
& 12.31 & 1.27 & 4.37 & 11.52 \\
\textbf{MedGemma-27B}
& 4.16 & 1.07 & 1.14 & 6.06
& 0.53 & 0.11 & 0.09 & 1.43
& 13.94 & 1.79 & 5.03 & 12.67 \\
\textbf{Qwen2.5-7B}
& 4.47 & 0.84 & 2.28 & 6.12
& 0.61 & 0.09 & 0.24 & 1.64
& 12.45 & 1.84 & 5.11 & 11.80 \\
\textbf{DeepSeek-VL2}
& 3.92 & 0.96 & 1.84 & 4.18
& 0.47 & 0.08 & 0.19 & 0.79
& 11.77 & 1.91 & 4.20 & 10.44 \\
\midrule
\textbf{Reweighting-All} & 16.71 & 10.29 & 7.15 & 33.50 & 2.19 & 0.65 & 1.98 & 6.42 & 14.02 & 7.42 & 14.37 & 37.84 \\
\textbf{Resampling-All} & 10.35 & 7.57 & 4.01 & 32.91 & 0.68 & 0.31 & 0.01 & 6.04 & 23.72 & 11.26 & 14.34 & 38.67 \\
\rowcolor{mygray}
\textbf{FairLLaVA-All}
& 8.93 & 6.65 & 3.45 & 30.53
& 0.01 & 0.82 & 0.04 & 4.63
& 25.11 & 12.03 & 15.66 & 40.03 \\
\bottomrule
\end{tabular}}
\caption{Fairness Gaps (First two main columns across Age, Gender) and Overall performance (last column) on PadChest dataset. Highlights tradeoff between Overall-Performance and Fairness-Gaps.}
\label{tab:fair_gaps_padchest}
\end{table*}

\section{Implementation Details}
\subsection{Preprocessing for HAM10000}
For HAM10000, QA data were generated using a concept-grounded synthesis pipeline built on both a language model and a vision--language model  using SelfSynthx~\cite{selfsynthx}. We first used OpenAI GPT-4o (\texttt{gpt-4o-2024-08-06}) to extract class-level dermoscopic concepts from the diagnosis labels (MEL, NV, BCC, AKIEC, BKL, DF, VASC), after mapping each code to its full clinical name to reduce ambiguity. This produced a label-to-concept bank of dermatology-relevant visual descriptors. We then generated image-level candidate descriptions using LLaVA-1.5-7B (\texttt{llava-1.5-7b-hf}, served via vLLM), and scored their relevance to the concept bank using \texttt{e5-large-v2} embeddings with an InfoNCE-style selection step to retain the most informative concepts for each image.

Using the selected concepts, we synthesized diverse question types, including short diagnostic, explanatory, and reasoning-style forms, and generated candidate answers with the same VLM, LLaVA-1.5-7B. To ensure correctness, candidate QA pairs were filtered for diagnosis consistency using exact label mention or fuzzy string matching against the ground-truth HAM10000 diagnosis; when consistency was weak, a conservative fallback answer was used. The final output for each image was a curated \texttt{new\_QA} set containing diagnosis-consistent, concept-supported question-answer pairs. Please refer to SelfSynthx~\cite{selfsynthx} for more details.

\subsection{Preprocessing on PadChest}
\begin{figure}[t]
    \centering
    \includegraphics[width=\linewidth]{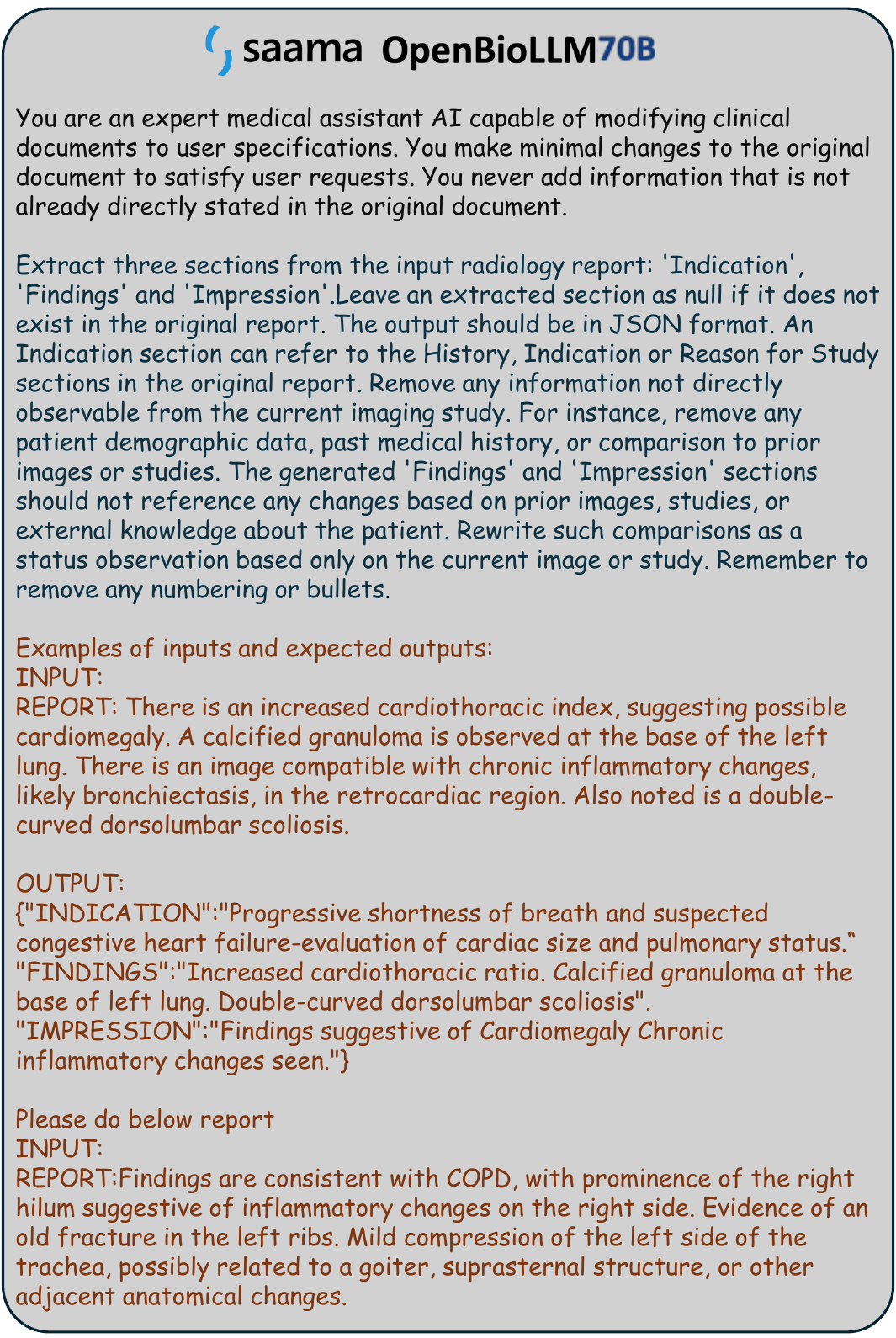}
    \caption{OpenBio-LLM Prompt used for preprocessing the PadChest Dataset to convert full report summary into standard radiology report section - "Findings", "Indication", "Impression".}
    \label{fig:prompt_image}
\end{figure}

For the English translated version of the PadChest~\cite{bustos2020padchest, stanford_rrg_challenge} dataset we follow ~\cite{llavarad} to pre-process the whole report summary into standard radiology report sections such as "Findings", "Indication" and "Impression". Indication section briefly states the reason why the study was ordered and the clinical question it aims to address (e.g., symptoms, suspected diagnosis). Findings section provides a description of what is observed in the images, without overall judgment. Impression includes interpretive summary of the key findings, likely diagnoses, and any critical recommendations. This work is focused on the producing the findings section of the radiology report. We also remove mentions of dates of previous studies each report references. We prompt~\cite{OpenBioLLMs-1} for pre-processing. Example of such prompt is given in~\cref{fig:prompt_image}.

\subsection{Implementation of Baselines}
For a fair comparison, all the reweighting, resampling, Adv. classifier baselines follow exactly similar instantiation of the Language Model and in-domain Image Embedder as FairLLaVA. We use Vicuna-7b-v1.5~\cite{vicuna2023} as our base language model and BioMedCLIP~\cite{biomedclip} as the image encoder.
Adv. MLP classifier uses same FairLLaVA DAC architecture of 2 layer MLP and uses middle three layers (14, 16, 20) for debiasing for one epoch. Before debiasing, Adv-MLP classifier is pretrained on LLaVA-Rad for three epochs achieving overall demography classification accuracy of $72\%$. All the baselines are either trained on 8 NVIDIA RTX-A6000 GPUs or 2 NVIDIA A100 with effective batch size of 96 for one epoch on all datasets. We use the same seed as used in FairLLaVA to control for randomness in weight initialization and otherwise in the implementation.

\section{Additional Ablations}
To characterize the effect of the proposed modules for demographic information minimization, we compare two training strategies for the DAC classifier when debiasing with respect to the \emph{Age} attribute on the MIMIC-CXR dataset:

\begin{enumerate}[leftmargin=*]
    \item \textbf{Pretrained DAC only.} We first pretrain the DAC classifier for three epochs using only the DAC loss $\mathcal{L}_{DAC}$ while keeping the base LLaVA-Rad model frozen. In the subsequent debiasing stage, we freeze this classifier and optimize the report generator with $\mathcal{L}_{LM} + \mathcal{L}_{DIM}$ using the fixed DAC as an adversary. This corresponds to the row \emph{FairLLaVA-Age ($\mathcal{L}_{DAC}$, then  $\mathcal{L}_{LM}+\mathcal{L}_{DIM}$)} in \cref{tab:mimic_ablation1_suppl}.
    \item \textbf{Joint DAC + MI training.} In the second setting, which we use in all main experiments, the DAC is \emph{not} pretrained. Instead, it is trained jointly with the report generator during fairness-aware fine-tuning, with the combined objective $\mathcal{L}_{LM}+\mathcal{L}_{DIM}+\mathcal{L}_{DAC}$. This corresponds to the row \emph{FairLLaVA-Age ($\mathcal{L}_{LM}+\mathcal{L}_{DIM}+\mathcal{L}_{DAC}$)}.
\end{enumerate}

All ablations are run for one debiasing epoch. As shown in \cref{tab:mimic_ablation1_suppl}, joint training of the DAC yields consistently better equity-scaled metrics \emph{and} higher overall performance. For the age-focused ES-metrics, ES-BLEU-4 improves from $2.48$ to $5.47$, ES-RadGraph-F1 from $3.50$ to $3.75$, and ES-GREEN from $2.08$ to $2.36$ when moving from the pretrained DAC to the jointly trained DAC. At the same time, overall report quality improves: BLEU-1 rises from $31.77$ to $35.28$, BLEU-4 from $10.48$ to $13.96$, RadGraph-F1 from $27.16$ to $28.49$, and GREEN from $33.06$ to $34.12$.

These results highlight a key limitation of using a strong pretrained attribute classifier as an adversary. Because the DAC is optimized in isolation and then frozen, its gradients during debiasing tend to aggressively remove age-related signal from the shared representation, including clinically relevant features, which leads to \emph{catastrophic forgetting} of some concepts and a noticeable drop in overall performance. This behavior is previously noted as limitation of Adv. MLP classifier~\cite{dear_CVPR}, where pretrained demographic classifiers can over-regularize the model. In contrast, jointly training the DAC with $\mathcal{L}_{DIM}$ and $\mathcal{L}_{LM}$ allows the model to gradually disentangle demographic information while being continually constrained by the primary reporting objective, resulting in a more balanced trade-off between equity-scaled performance and overall clinical utility.

\begin{table*}[h!]
\centering
\begingroup
\setlength{\tabcolsep}{3pt}   
\renewcommand{\arraystretch}{0.95} 

\newcommand{\rt}[1]{\rotatebox[origin=c]{45}{\makebox[1.6cm][c]{\scriptsize\bfseries #1}}}

\resizebox{0.95\linewidth}{!}{
\begin{tabular}{l*{8}{c}}
\toprule
\multirow{2}{*}{\textbf{Method}} &
\multicolumn{4}{c}{\textbf{Age Group}\,$\uparrow$} &
\multicolumn{4}{c}{\textbf{Overall}\,$\uparrow$} \\
\cmidrule(lr){2-5}\cmidrule(lr){6-9}
& \rt{ES-BLEU-1} & \rt{ES-BLEU-4} & \rt{ES-RadGraph-F1} & \rt{ES-GREEN}
& \rt{BLEU-1} & \rt{BLEU-4} & \rt{RadGraph-F1} & \rt{GREEN} \\
\midrule


\textbf{FairLLaVA-Age} ($\mathcal{L}_{DAC}$ then$\mathcal{L}_{LM} + \mathcal{L}_{DIM}$)
& 3.83 & 2.48 & 3.50 & 2.08
& 31.77 & 10.48 & 27.16 & 33.06 \\

\textbf{FairLLaVA-Age} ($\mathcal{L}_{LM} + \mathcal{L}_{DIM} + \mathcal{L}_{DAC}$)
& \textbf{17.82} & \textbf{5.47} & \textbf{3.75} & \textbf{2.36}
& \textbf{35.28} & \textbf{13.96} & \textbf{28.49} & \textbf{34.12} \\
\bottomrule
\end{tabular}}
\endgroup
\caption{ES-M metrics for the Age Group attribute on the MIMIC-CXR dataset and overall performance. Joint training of the $\mathcal{L}_{DAC}$ with $\mathcal{L}_{DIM}$ improves equity-scaled scores while preserving or improving overall report quality.}
\label{tab:mimic_ablation1_suppl}
\end{table*}

\section{Additional Qualitative Samples}
\begin{figure*}[th!]
    \centering
    \includegraphics[width=\textwidth]{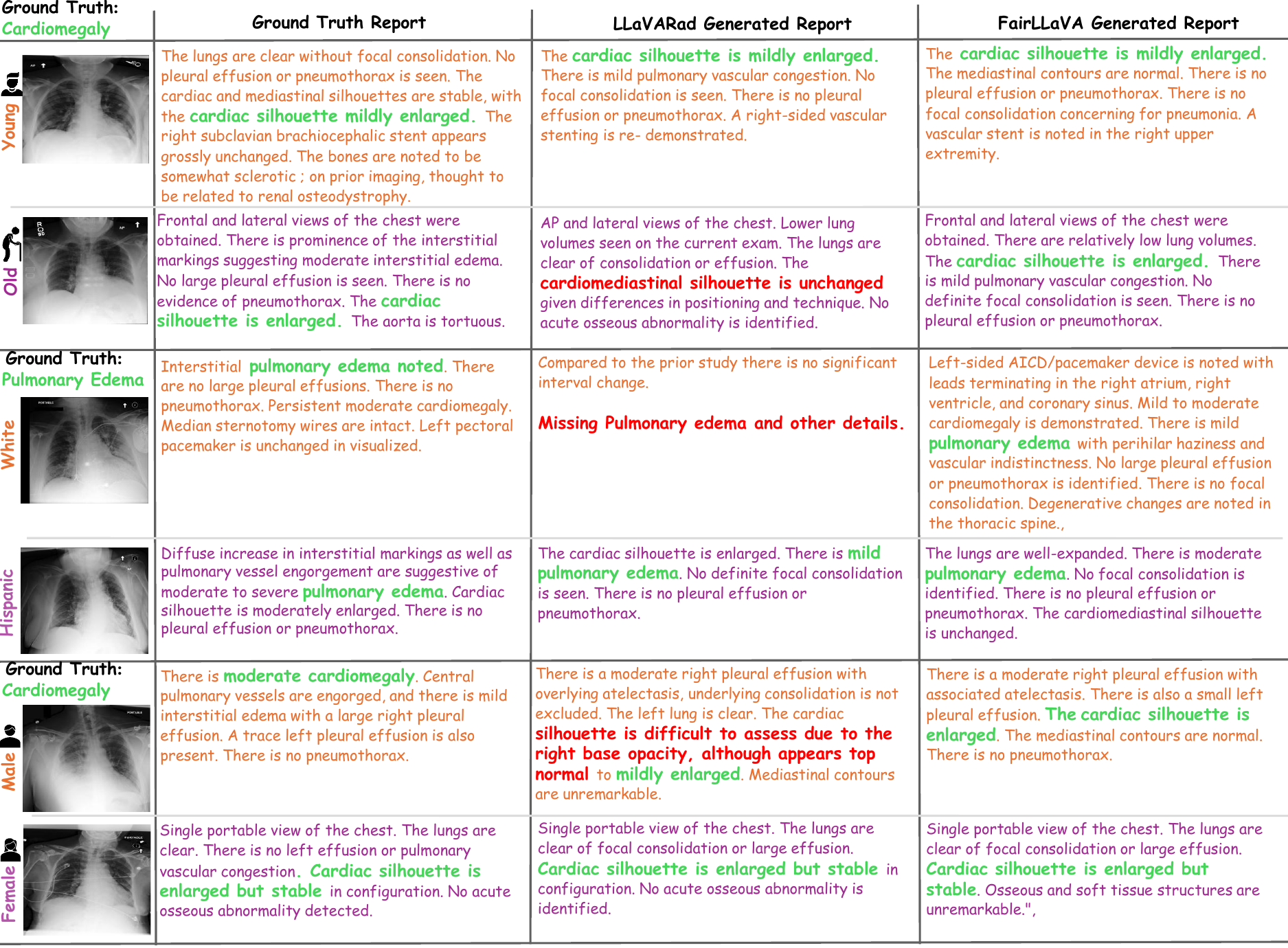}
    \caption{Qualitative samples show LLaVA-Rad baseline produces unfair results for some subgroups ("Old", “White”, "Male") of demographic attributes.} 
    \label{fig:qual_suppl1}
\end{figure*}

In~\cref{fig:qual_suppl1}, we show qualitative examples where the LLaVA-Rad baseline correctly identifies a key condition for one demographic subgroup but fails to mention the same finding for another, suggesting a dependence on spurious correlations with subgroup membership. In contrast, FairLLaVA, trained with our debiasing objectives, more consistently recovers the clinically relevant findings across all subgroups by focusing on image evidence rather than demographic cues. In some cases, such as the White–vs–Black pair, the baseline also appears less confident in its descriptions (Gender: Male in~\cref{fig:qual_suppl1}), whereas FairLLaVA provides clearer and more definitive statements about the underlying pathology in~\cref{fig:qual_suppl1}.

{
    \small
    \bibliographystyle{ieeenat_fullname}
    \bibliography{main}
}